\documentclass[10pt,twocolumn,letterpaper]{article}

\usepackage{iccv}              

\definecolor{iccvblue}{rgb}{0.21,0.49,0.74}
\usepackage[pagebackref,breaklinks,colorlinks,allcolors=iccvblue]{hyperref}
\usepackage{color, colortbl}
\usepackage{multirow}
\usepackage{adjustbox}

\definecolor{LightBlue}{RGB}{212, 250, 252} 
\definecolor{LightGreen}{RGB}{217, 250, 226}

\def\paperID{14} 
\def\confName{ICCV}
\def\confYear{2025}

\usepackage{tikz}
\newcommand\copyrighttext{%
  \footnotesize \textcopyright 2025 IEEE.  Personal use of this material is permitted.  Permission from IEEE must be obtained for all other uses, in any current or future media, including reprinting/republishing this material for advertising or promotional purposes, creating new collective works, for resale or redistribution to servers or lists, or reuse of any copyrighted component of this work in other works.}
\newcommand\copyrightnotice{%
\begin{tikzpicture}[remember picture,overlay]
\node[anchor=south,yshift=10pt] at (current page.south) {\fbox{\parbox{\dimexpr\textwidth-\fboxsep-\fboxrule\relax}{\copyrighttext}}};
\end{tikzpicture}%
}

\title{Extracting Uncertainty Estimates from Mixtures of Experts for Semantic Segmentation}

\author{Svetlana Pavlitska$^{1,2}$, Beyza Keskin$^{1}$, Alwin Faßbender$^{1}$, Christian Hubschneider$^{2}$, J. Marius Zöllner$^{1,2}$\\
\textit{$^{1}$ Karlsruhe Institute of Technology (KIT), Germany}\\
\textit{$^{2}$ FZI Research Center for Information Technology, Germany} \\
{\tt\small pavlitska@fzi.de}\\
}

\begin{document}
\maketitle
\copyrightnotice
\thispagestyle{empty}
\pagestyle{empty}

\begin{abstract}
Estimating accurate and well-calibrated predictive uncertainty is important for enhancing the reliability of computer vision models, especially in safety-critical applications like traffic scene perception. While ensemble methods are commonly used to quantify uncertainty by combining multiple models, a mixture of experts (MoE) offers an efficient alternative by leveraging a gating network to dynamically weight expert predictions based on the input. Building on the promising use of MoEs for semantic segmentation in our previous works, we show that well-calibrated predictive uncertainty estimates can be extracted from MoEs without architectural modifications. We investigate three methods to extract predictive uncertainty estimates: predictive entropy, mutual information, and expert variance. We evaluate these methods for an MoE with two experts trained on a semantical split of the A2D2 dataset. Our results show that MoEs yield more reliable uncertainty estimates than ensembles in terms of conditional correctness metrics under out-of-distribution (OOD) data.
Additionally, we evaluate routing uncertainty computed via gate entropy and find that simple gating mechanisms lead to better calibration of routing uncertainty estimates than more complex classwise gates. Finally, our experiments on the Cityscapes dataset suggest that increasing the number of experts can further enhance uncertainty calibration. Our code is available at \url{https://github.com/KASTEL-MobilityLab/mixtures-of-experts/}.
\end{abstract}

\section{Introduction}
In a mixture of experts (MoE), a complex task is divided among multiple specialized models (experts), and a gating network is used to dynamically select or weigh their contributions based on the input~\cite{jacobs1991adaptive}. While in an ensemble model predictions are combined deterministically using averaging or voting, an MoE provides input-dependent weighting of the predictions provided by the expert models.

Depending on the implementation, MoEs in deep learning can operate at the model level, where entire neural networks serve as experts, or at the layer level, where components of a network act as experts. Layer-level MoEs gained popularity in natural language processing~\cite{shazeer2017outrageously,rajbhandari2022deepspeed,fedus2022switch,huang2024harder} and computer vision~\cite{wang2019deep,riquelme2021scaling,pavlitska2023sparsely} for their ability to significantly increase model capacity without proportionally increasing inference time. Model-level MoEs are commonly applied in computer vision~\cite{ahmed2016network,pavlitskaya2020using,he2021deepme}.

\begin{figure}[t]
\centering
\begin{subfigure}[t]{0.24\columnwidth}
        \centering
        \includegraphics[width=\textwidth]{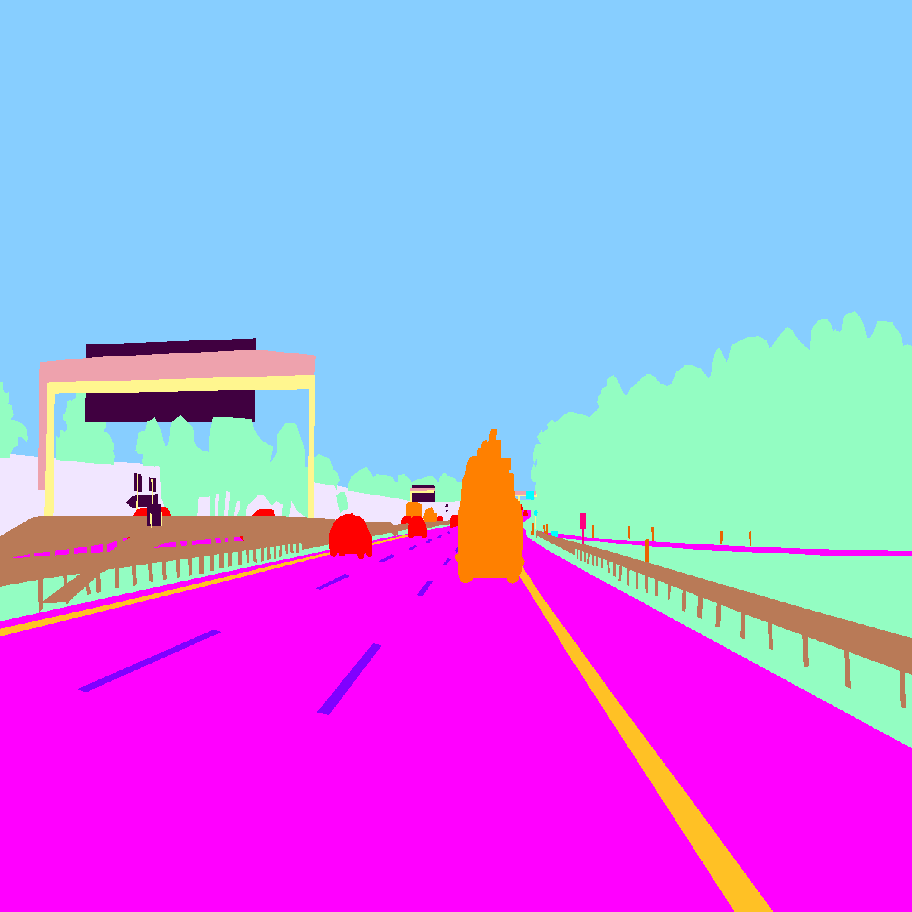}
        \caption*{Ground truth}
    \end{subfigure}
     \begin{subfigure}[t]{0.24\columnwidth}
        \centering
        \includegraphics[width=\textwidth]{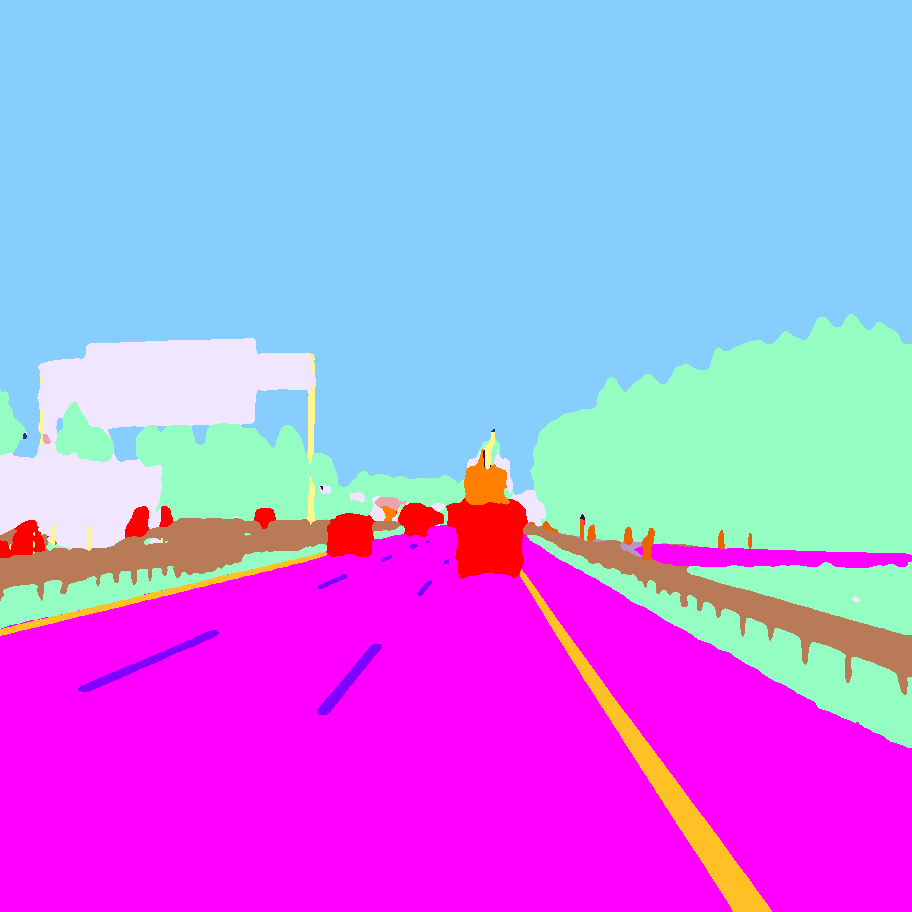}
        \caption*{MoE prediction}
    \end{subfigure}
    \begin{subfigure}[t]{0.24\columnwidth}
        \centering
        \includegraphics[width=\textwidth]{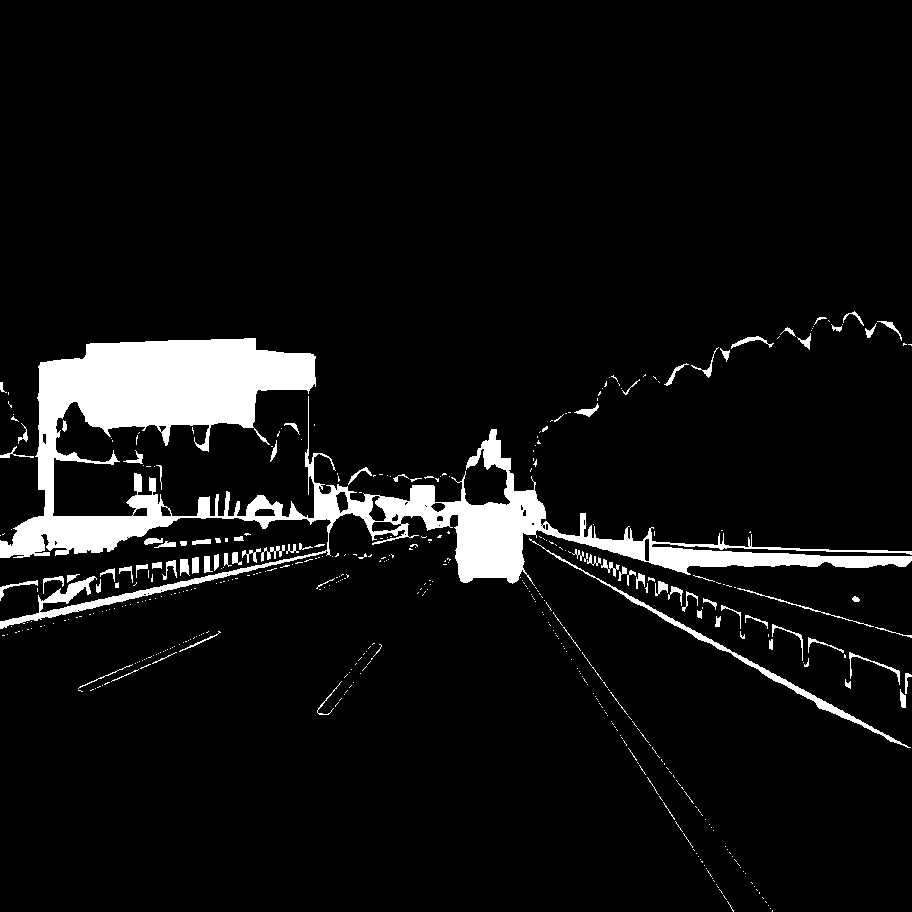}
        \caption*{MoE inaccuracy}
    \end{subfigure}
    \begin{subfigure}[t]{0.24\columnwidth}
        \centering
        \includegraphics[width=\textwidth]{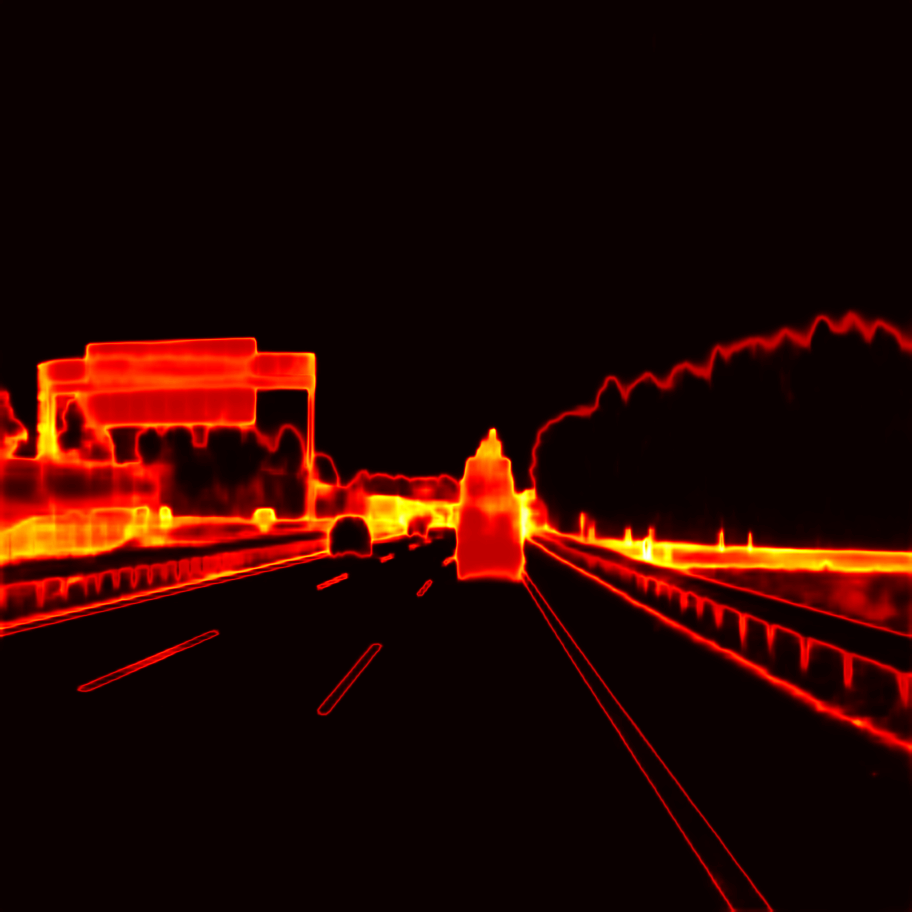}
        \caption*{MoE uncertainty}
    \end{subfigure}
    \caption{We extract uncertainty estimates from an MoE combining pre-trained expert models for semantic segmentation. }
    \label{fig:concept}
\end{figure}


Disagreements between the outputs of the experts and the resulting MoE were shown to help identify regions of an input image, which are challenging for the semantic segmentation models~\cite{pavlitskaya2020using}. Compared to this disagreement analysis, uncertainty quantification can provide a more systematic and probabilistic understanding of model confidence. In particular, uncertainty estimation methods yield quantifiable measures, allow for calibration of model confidence~\cite{blundell2015weight}, enable applications such as active learning~\cite{didari2024bauesian}, and help to distinguish between epistemic (i.e., arising from the lack of knowledge or data) and aleatoric (i.e., arising from inherent noise in data) uncertainties~\cite{kendall2017what}.

Previous works addressed incorporating uncertainty consideration directly into MoEs~\cite{luttner2023training,zhang2023efficient}. In this work, we extract uncertainty from MoEs without intentional architectural changes and conduct experiments on mixtures of pre-trained experts. These models have already demonstrated semantic segmentation performance superior to single models and ensembles~\cite{pavlitskaya2020using,pavlitskaya2022evaluating}. We show that extracted uncertainty estimates outperform ensembles, especially on the out-of-distribution (OOD) data and under data shift. 






\section{Related Work}

Uncertainty is the lack of certainty about a model's prediction, divided into \textbf{aleatoric} (due to intrinsic randomness in data) and \textbf{epistemic uncertainty} (due to limited knowledge or model limitations)~\cite{kendall2017what}. While aleatoric uncertainty cannot be eliminated (but can be mitigated by data augmentation or redundant input), the epistemic one can be reduced by enlarging the amount or diversity of training data or increasing model capacity, e.g., with ensembles or MoEs.

\subsection{Uncertainty Quantification for Semantic Segmentation}

Many existing works on uncertainty estimation focus on image classification~\cite{gal2016dropout,lakshminarayanan2017simple,blundell2015weight}. Semantic segmentation is inherently more complex, so uncertainty quantification methods require modifications to perform estimation per pixel, resulting in spatial uncertainty maps instead of a single scalar uncertainty score as for classification tasks.

Aleatoric uncertainty can be captured for in-distribution samples in a single forward pass via the entropy of the softmax predictions~\cite{mukhoti2018evaluating}, computed per pixel for semantic segmentation. However, choosing a loss function, network depth and architecture, poor regularization, and calibration issues can lead to overconfident predictions, making softmax entropy an unreliable uncertainty measure. Regularization~\cite{pereyra2017regularizing}, temperature scaling~\cite{guo2017calibration}, label smoothing~\cite{szegedy2016rethinking}, and alternative losses can mitigate the problem of overconfidence. In particular, post-hoc calibration methods like temperature scaling can adjust the predicted probabilities of a trained model to better reflect true confidence in predictions without retraining. These methods are computationally cheap but limited in flexibility; they do not adjust the underlying feature representations and are sensitive to calibration data, which should be available as a hold-out set.

More robust uncertainty estimates can be captured using uncertainty-aware training methods like ensembles or Bayesian approaches. Deep ensembles assess uncertainty estimation based on the output variance of pre-trained member networks~\cite{lakshminarayanan2017simple}. Uncertainty stemming from ensembles is more reliable; it surpasses post-hoc calibration methods and Bayesian neural networks~\cite{ovadia2019can}. However, the disadvantage is a significant computational overhead for training and inference of several models. Ensembles also improved uncertainty estimation for semantic segmentation under domain shift~\cite{cygert2021closer}. Finally, Monte Carlo (MC) Dropout~\cite{gal2016dropout} can be applied to generate a posterior distribution of pixel class labels for semantic segmentation~\cite{kendall2017bayesian}.

\subsection{Uncertainty Quantification via MoEs}
Model-level MoEs can be used for uncertainty estimation by leveraging expert disagreement, gating confidence, or specialized uncertainty-aware experts. Jiang et al.~\cite{jiang2023mixture} introduced an MoE approach that leverages uncertainty estimation to address data imbalance in regression tasks. The model predicts sample-wise aleatoric uncertainty and utilizes it for expert fusion, leading to improved performance and better-calibrated uncertainty estimates. Another approach to intentionally introduce uncertainty quantification in the model architecture to capture aleatoric uncertainty for semantic segmentation tasks was proposed by Gao et al.~\cite{gao2023modeling}. In a mixture of stochastic experts (MoSE), each expert estimates a distinct mode of uncertainty, and a gate predicts the probabilities of an input image being segmented in those modes, resulting in a two-level uncertainty representation. Finally, Cao et al.~\cite{cao2022uncertainty} proposed a multi-gating approach for uncertainty calibration, leading to better calibration than the MC dropout baseline. The evaluation was performed for the time series regression task.


In addition to model-level approaches, layer-level MoEs can be used for uncertainty estimation, particularly in cases where computational efficiency is crucial. 
Layer-level MoEs were shown to capture epistemic and aleatoric uncertainty through the selective activation of experts. Zheng et al.~\cite{zheng2019self} proposed a self-supervised MoE framework for multitask reinforcement learning driven by predictive uncertainty estimation. The model employs a gate calibrated by expert uncertainty feedback, improving sample efficiency and generalization across multiple tasks. Luttner~\cite{luttner2023training} proposed the uncertainty-aware Mixture of Experts (uMoE), designed to handle aleatoric uncertainty during the training phase. By partitioning the uncertain input space and training specialized experts, the model dynamically adjusts to minimize deviations from ground truth, enhancing robustness in various data-driven domains. Zhang et al.~\cite{zhang2023efficient} proposed using an uncertainty-aware gate for routing in a visual transformer for the weather removal task. It used MC dropout for uncertainty estimation to weigh features and route them to the experts correspondingly. The proposed routing was shown to enhance expert specialization via better task-specific routing. 

Our work focuses on model-level MoEs, as their potential for disagreement analysis has already been shown~\cite{pavlitskaya2020using}. Unlike most related works, however, we do not introduce architectural changes in an MoE to extract uncertainty. Instead, we adapt the uncertainty extraction methods proposed initially for ensembles. Avoiding architectural changes to MoEs preserves their efficiency and compatibility with existing pretrained models, enabling uncertainty estimation without increasing implementation complexity or retraining overhead. 

\clearpage
\newpage
\section{Method}
This section describes the methods to extract and measure uncertainties from model-level MoEs.

\subsection{Model-level MoE}

A model-level MoE combines the outputs of \( N \) independently trained expert models using a gate that assigns input-dependent weights to each expert. For a given input \( x \) and class \( c \), the final class probability predicted by the MoE is computed as a weighted sum of the expert outputs:

\begin{equation}
p_{\text{MoE}}(c \mid x) = \sum_{e=1}^N w_e(x) \cdot p_e(c \mid x),
\end{equation}

where:
\begin{itemize}
    \item \( p_e(c \mid x) \) is expert \( e \)'s probability of predicting class \( c \),
    \item \( w_e(x) \) is the gate weight assigned to expert \( e \) for input \( x \).
\end{itemize}

\subsection{Predictive Uncertainty Estimates in MoEs}

\textbf{Predictive entropy (PE)} is typically computed by averaging the output probability distributions from multiple stochastic forward passes (e.g., using MC Dropout~\cite{blundell2015weight} or ensembles~\cite{lakshminarayanan2017simple}), and then measuring the Shannon entropy of the resulting aggregated predictive distribution $\bar{p}(c \mid x)$ to capture both aleatoric and epistemic uncertainty of the model's predictions: $\text{PE}(x) = H(\bar{p}(c \mid x))$.

\textbf{Mutual Information (MI)} quantifies epistemic uncertainty arising from uncertainty in the model parameters. MI is computed by measuring the difference between the uncertainty of the average prediction (i.e., the PE) and the average uncertainty of the individual model predictions~\cite{houlsby2011bayesian}. MI captures epistemic uncertainty, reflecting uncertainty due to limited model knowledge or expert disagreements.

For an ensemble, the aggregated predictive distribution $ \bar{p}(c \mid x) $ is typically computed as an average over model outputs, i.e., $\bar{p}(c \mid x) = \frac{1}{N} \sum_{e=1}^N p_e(c \mid x)$. For an MoE, this computation would ignore the gating and thus be analogous to using an ensemble. To address this, we propose two approaches to compute $ \bar{p}(c \mid x) $ for PE and MI. 

In the \textbf{stacked approach to compute $ \bar{p}(c \mid x) $ for PE and MI}, the MoE output is added as a further expert output:

\begin{equation}
  \bar{p}_{\text{stacked}}(c \mid x) = \frac{1}{N + 1} \left(p_{\text{MoE}}(c \mid x) + \sum_{e=1}^N p_e(c \mid x) \right)
\end{equation}
Here, we include the MoE behavior to improve alignment with the final prediction. Since the MoE output is not independent of the experts, including it may, however,  artificially reduce perceived variance. 

In the \textbf{weighted approach to compute $ \bar{p}(c \mid x) $ for PE and MI}, only expert outputs are taken, multiplied by the gate weights: the aggregated predictive distribution is thus equivalent to the MoE output itself. This approach explicitly incorporates the gating logic and is therefore better aligned with the MoE's internal decision logic. However, it assumes that the gate returns a single weight value for each expert, which is not true in some architectures.

In addition to PE and MI, we propose \textbf{expert variance (EV)} as a further uncertainty extraction method based on the variability of expert predictions within the MoE, following the idea of using standard deviation maps to quantify uncertainty~\cite{holder2021efficient}. EV for a given input $x$ and class $c$ is defined as the pixel-wise variance of each expert's predictions relative to the final MoE output:

\begin{equation}
\text{EV}(x, c) = \frac{1}{N} \sum_{e=1}^N \left(p_e(c \mid x) - p_{\text{MoE}}(c \mid x)\right)^2
\end{equation}

EV captures epistemic and aleatoric uncertainty, reflecting how much expert predictions diverge from MoE output. 

\subsection{Routing Uncertainty Estimates in MoEs}
In addition to the predictive uncertainty captured by PE, MI, and partially by EV, the \textbf{routing uncertainty} can also be extracted from an MoE, reflecting how confidently the gate selects among experts. To quantify it, we use \textbf{gate entropy}, measuring the uncertainty of the gating distribution over experts. A low gate entropy indicates confident, focused routing to a specific expert, while high entropy suggests that the MoE is indecisive or diffuse in assigning responsibility. Gate entropy can thus serve as a further signal about the model's internal decision-making process.

\subsection{Uncertainty Evaluation Metrics}

Calibration reflects the alignment between predicted confidence and observed accuracy~\cite{lakshminarayanan2017simple,ovadia2019can,hubschneider2019calibrating}. We use expected calibration error (ECE)~\cite{nixon2019measuring,ovadia2019can}, maximum calibration error (MCE)~\cite{naeini2016obtaining,ovadia2019can}, Brier score~\cite{brier1950verification}, and negative log-likelihood (NLL). ECE and MCE evaluate the gap between predicted confidence and observed accuracy across bins of confidence levels. Brier Score and NLL assess the probabilistic accuracy of predictions. Lower values indicate better calibration, reflecting higher confidence in correct predictions and penalizing overconfident incorrect ones. 

We also use the conditional correctness metrics $p$(accurate$\vert$certain), $p$(uncertain$\vert$inaccurate), and Patch Accuracy vs Patch Uncertainty (PAvPU) proposed by Mukhoti and Gal~\cite{mukhoti2018evaluating} for semantic segmentation. They measure the ability to align high confidence with correct predictions and uncertainty with incorrect predictions, respectively. For these metrics, the image is divided into patches, and each patch's accuracy and uncertainty are evaluated. Patches with accuracy above a threshold are labeled as accurate, and those with uncertainty above a threshold are labeled as uncertain. 

\clearpage
\newpage
\section{Evaluation of MoEs with Semantically Disjoint Experts}
\label{sec:disjoint}
\begin{figure}[t]
    \centering
        \includegraphics[width=\linewidth]{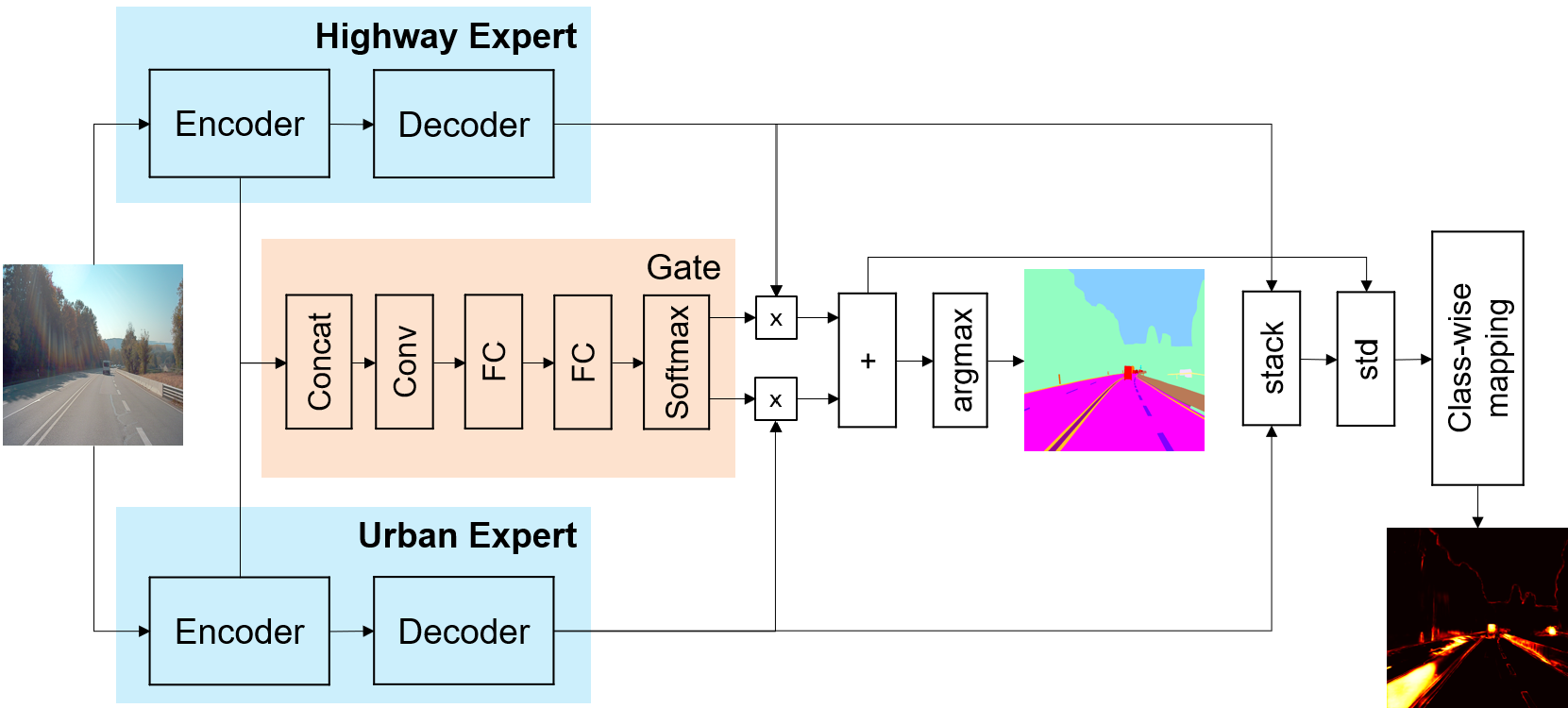}
    \caption{MoE architecture with two experts and an additional convolutional layer.}
    \label{fig:moe-archl}
\end{figure}

In the first part of the evaluation, we focus on an MoE architecture for semantic segmentation of traffic scenes from our previous works~\cite{pavlitskaya2020using,pavlitskaya2022evaluating} (see Figure~\ref{fig:moe-archl}) comprising two experts, trained on semantically disjoint data subsets: urban and highway traffic scenes. 

\subsection{Experimental Setup}

\textbf{Models}: The experts architecture is \texttt{DeepLabv3+}~\cite{chen2018encoder} with a \texttt{ResNet-101} backbone~\cite{he2016deep}, pre-trained on ImageNet~\cite{russakovsky2015imagenet}. In an MoE, the features are extracted from each expert's atrous spatial pyramid pooling layer (ASPP). The gate consists of one convolutional layer and two fully-connected layers. We use two gate architectures from~\cite{pavlitskaya2020using}: \textit{the simple gate} predicts a weight for each expert
, and \textit{the classwise gate} assigns weights for each class and expert
. If an additional convolutional layer is added after combining experts' predictions, the architecture is denoted as \textit{+ conv}.

\textbf{Dataset}: Evaluation was performed on the A2D2 dataset~\cite{geyer2020a2d2}. We used the split by road type proposed in~\cite{pavlitskaya2020using}: \textit{highway} and \textit{urban} subsets, each with 6,132 training, 876 validation, and 1,421 test samples, and an additional \textit{ambiguous} subset only with 1,421 test images.  

\textbf{Training:} An MoE contains an expert trained on the \textit{urban} and an expert trained on the \textit{highway} subset of data. To train an MoE, the experts are frozen, and only the gate weights and the weights of a final convolutional layer (if applicable) are trained on the combined \textit{highway-urban} dataset to learn the contribution of each expert based on input features. Baseline is a model with the same architecture as the exerts but trained on the combined \textit{highway-urban} dataset, i.e., it sees the same images as an MoE during training. We used trained models from our previous work~\cite{pavlitska2024towards}. 
Evaluation was performed on a combined \textit{highway-urban} test set with 2,842 images and on the \textit{ambiguous} test set with 1,421 images. In our setup, in-distribution (ID) refers to the highway and urban images used for training. In contrast, out-of-distribution (OOD) refers to images from the ambiguous subset containing scenes absent in the training data.



\subsection{Comparison Models}
We compare the MoE-based uncertainty estimation to two further methods: ensembles and MC dropout. 

\textbf{Ensemble}: The ensemble combines predictions of a highway and an urban expert via averaging. The softmax distributions \( p(y|x, \hat{\omega}) \) are obtained from the forward passes and are then used to calculate the PE and MI~\cite{lakshminarayanan2017simple}. Note that the used ensemble thus differs from deep ensembles as defined by Lakshminarayanan et al.~\cite{lakshminarayanan2017simple}, where ensemble members are trained on the same data but with random initialization of initial neural network parameters. 

\textbf{MC dropout}: We apply dropout layers during inference~\cite{gal2016dropout}. To estimate uncertainty, multiple stochastic forward passes are performed with dropout activated, resulting in a softmax distribution \( p(y|x, \hat{\omega}) \). Equivalent to an ensemble, PE is computed from the softmax distribution. In our experiments, we performed two forward passes for a fair comparison with an ensemble or MoE consisting of two models. We have experimented with a different number of forward passes (up to 20) and have not observed any improvement in the calibration metrics.

\begin{figure}[t]
    \centering
    \includegraphics[width=\linewidth]{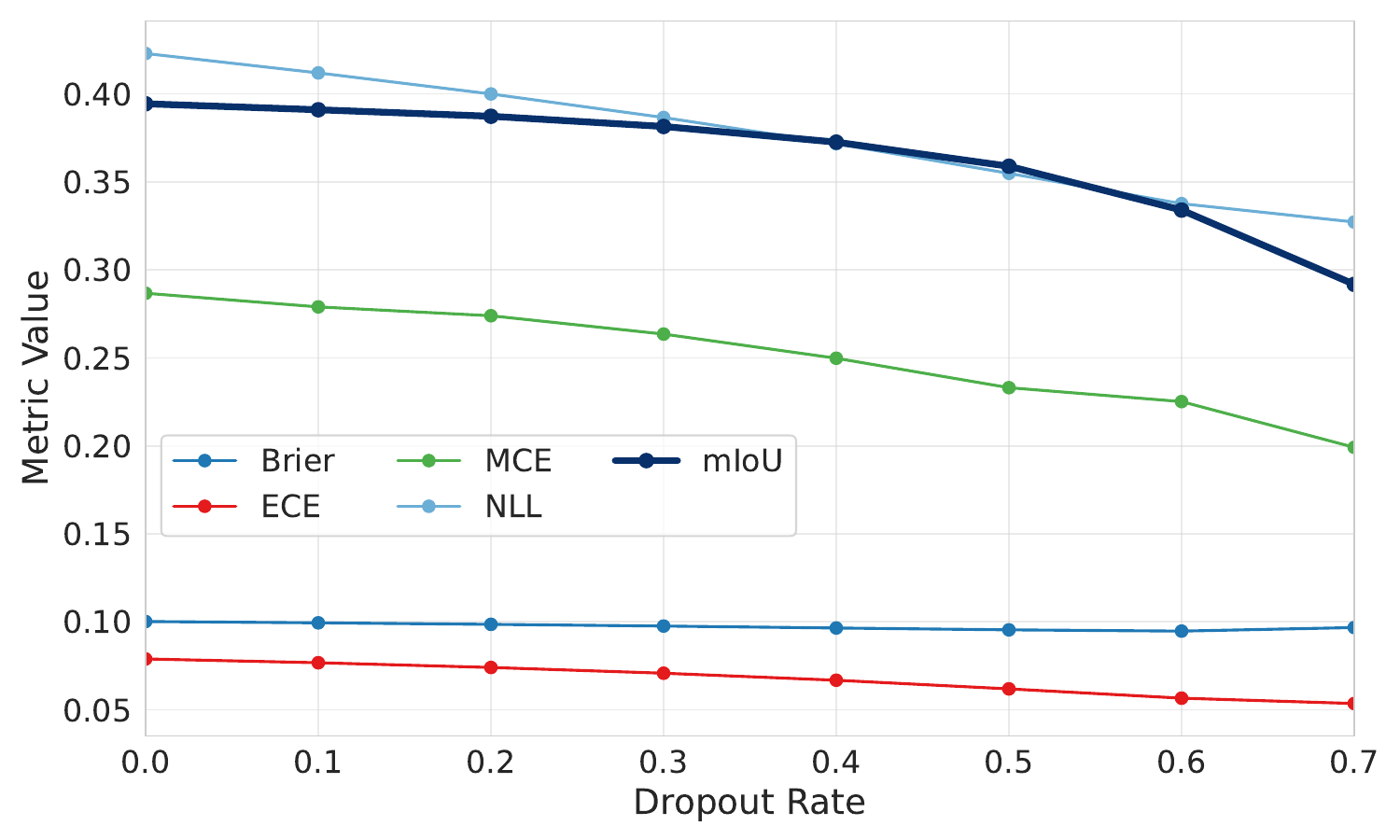}
    \caption{Semantic segmentation accuracy (dark blue) and uncertainty quantification via PE for the baseline with MC dropout.}
    \label{fig:dropout-rate}
\end{figure}

\begin{figure*}[h]
     \centering
     \includegraphics[width=0.6\linewidth]{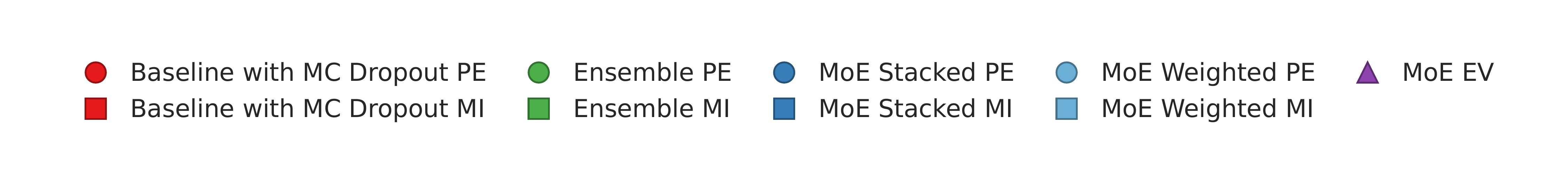}
     
     \begin{subfigure}{0.246\linewidth}
      \centering
      \includegraphics[width=\linewidth]{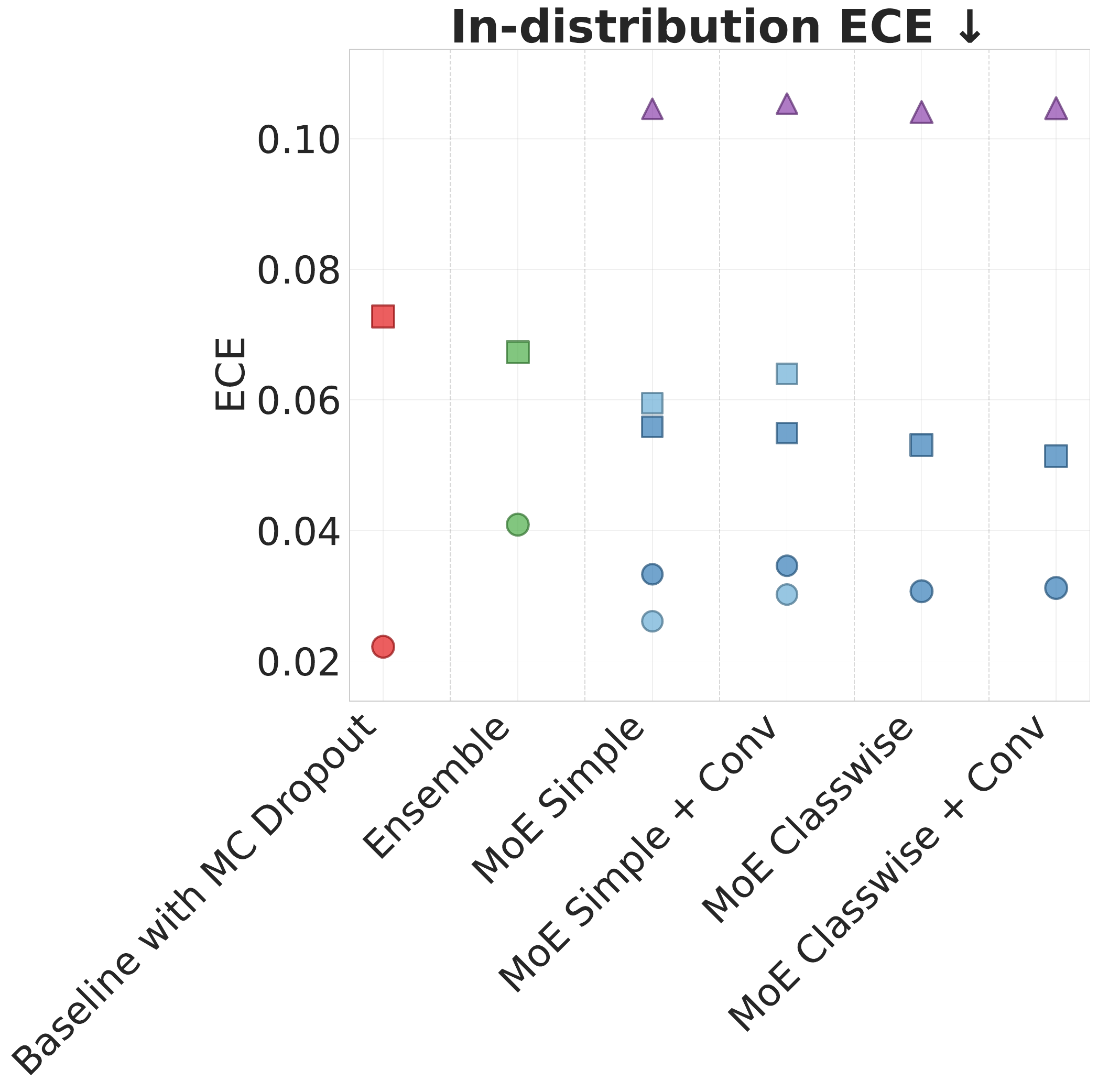}
    \end{subfigure}
    \begin{subfigure}{0.246\linewidth}
      \centering
      \includegraphics[width=\linewidth]{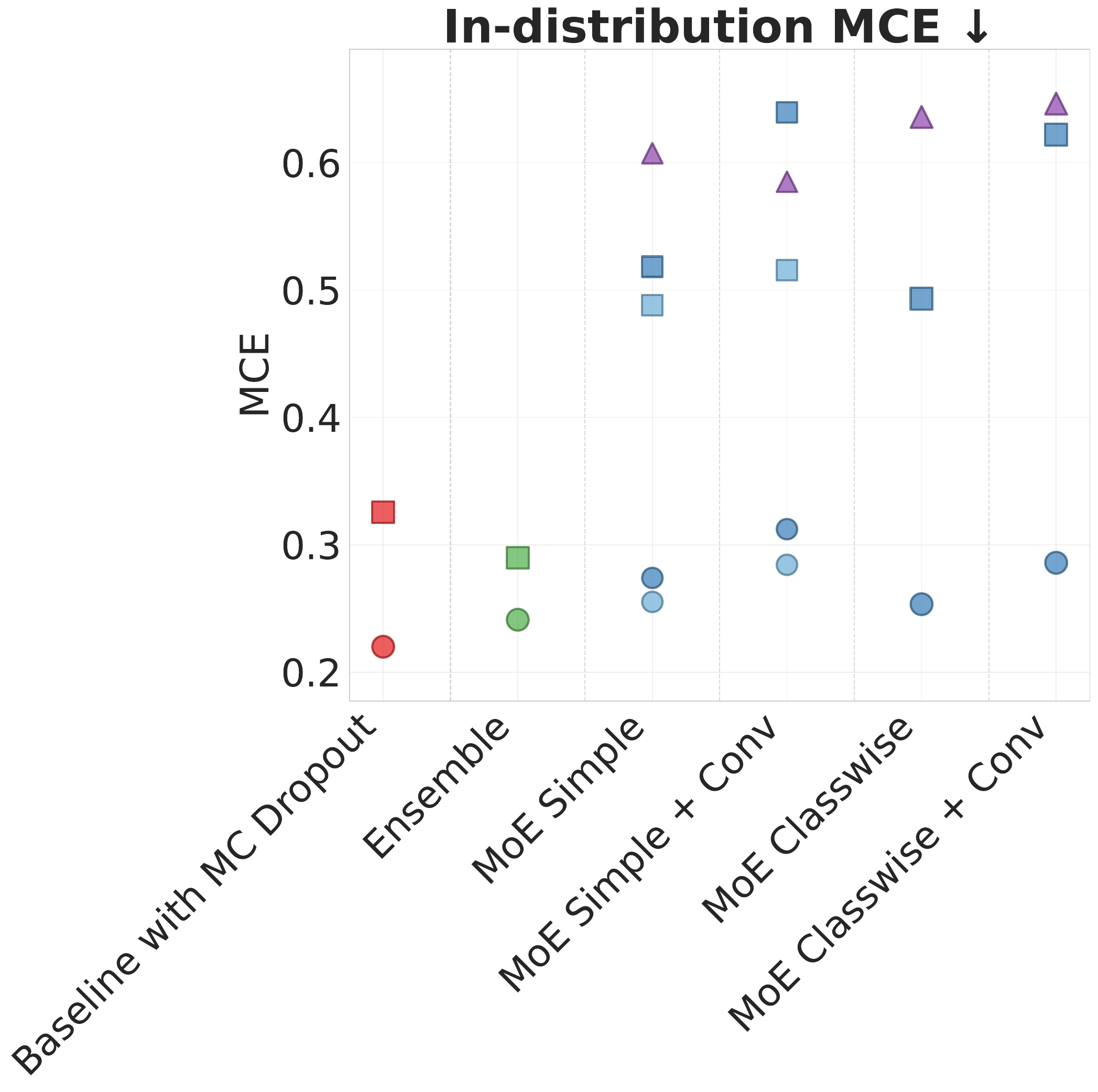}
    \end{subfigure}
    \begin{subfigure}{0.246\linewidth}
      \centering
      \includegraphics[width=\linewidth]{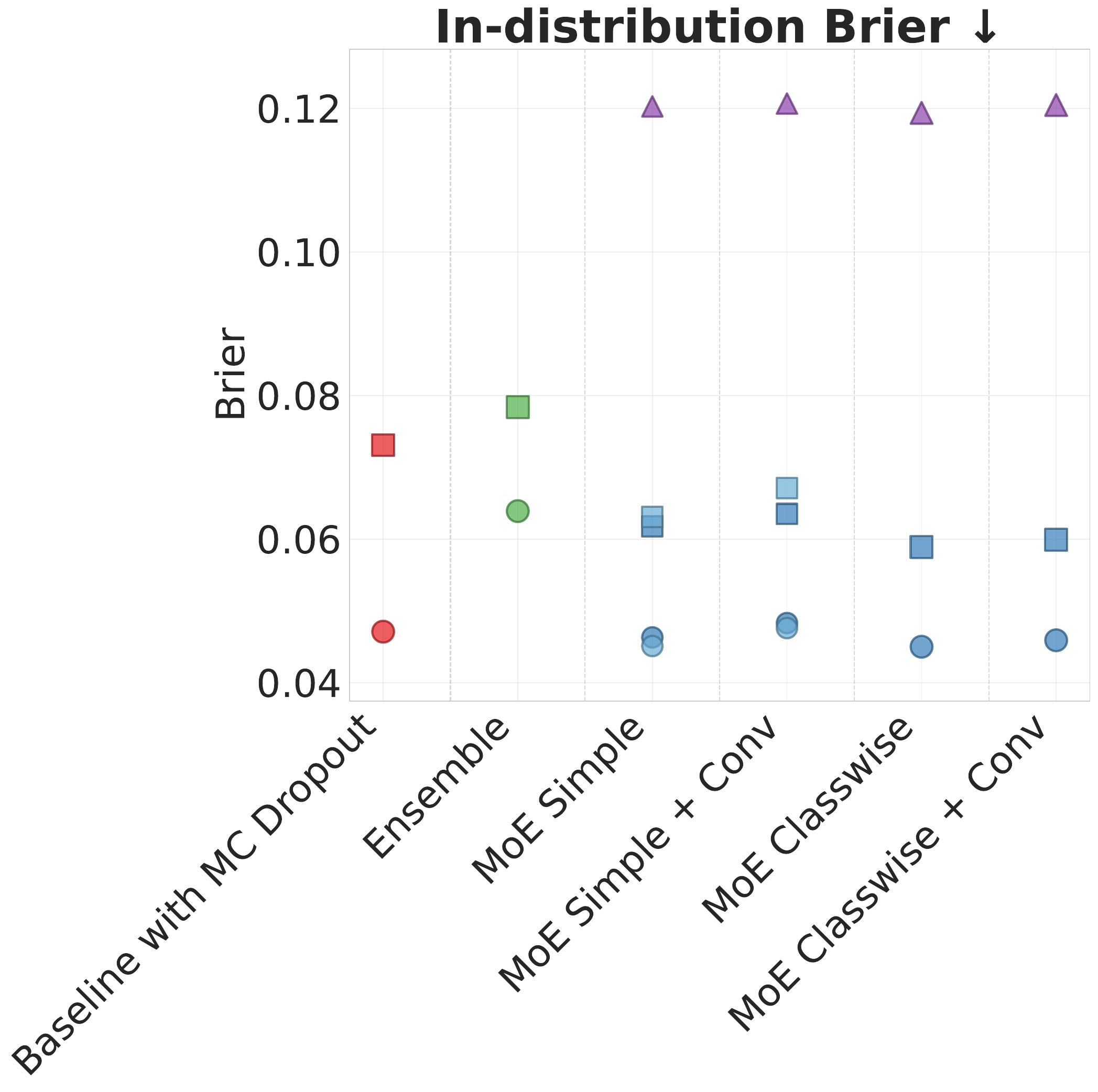}
    \end{subfigure}
    \begin{subfigure}{0.246\linewidth}
      \centering
      \includegraphics[width=\linewidth]{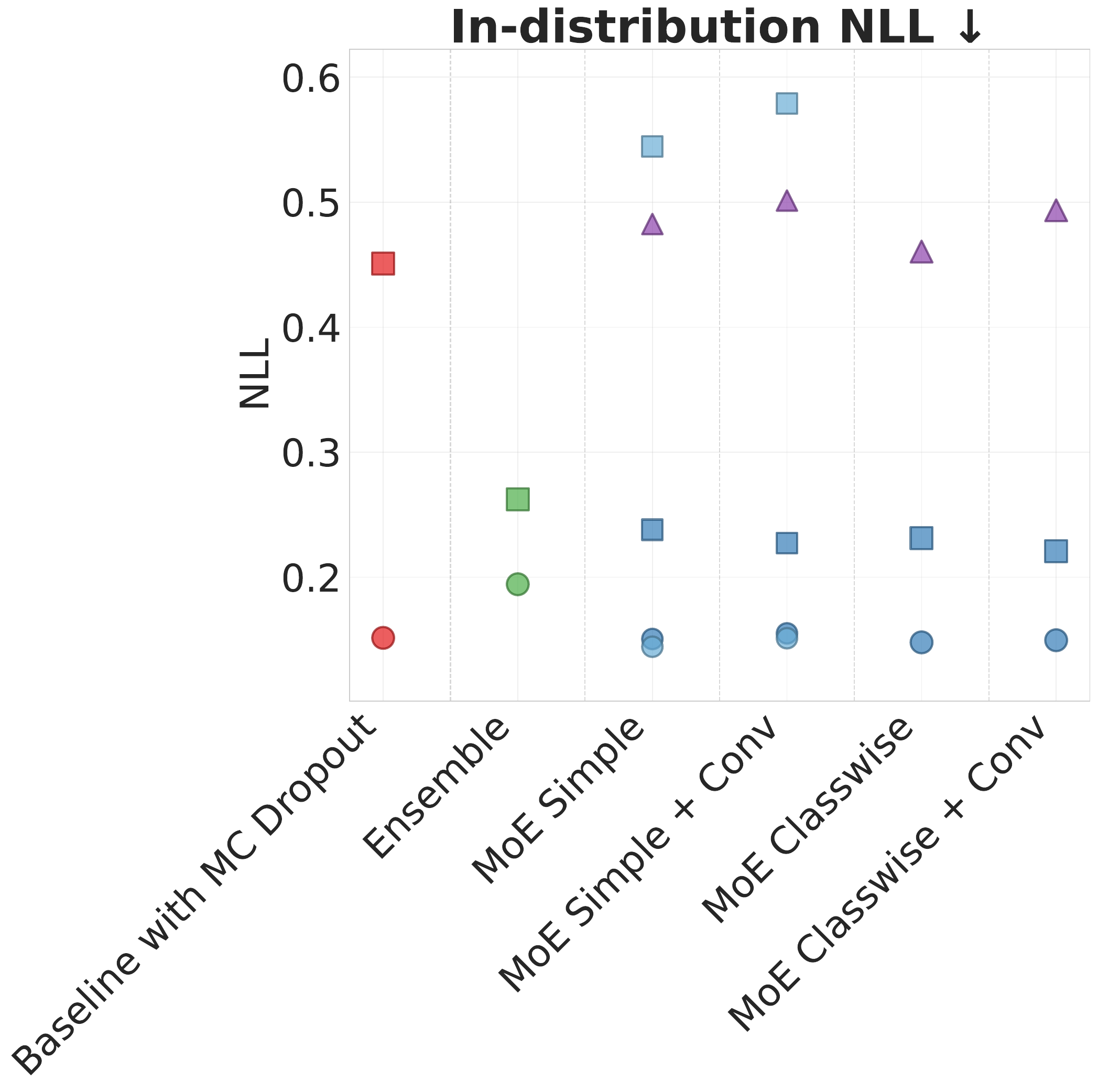}
    \end{subfigure}

     \begin{subfigure}{0.246\linewidth}
      \centering
      \includegraphics[width=\linewidth]{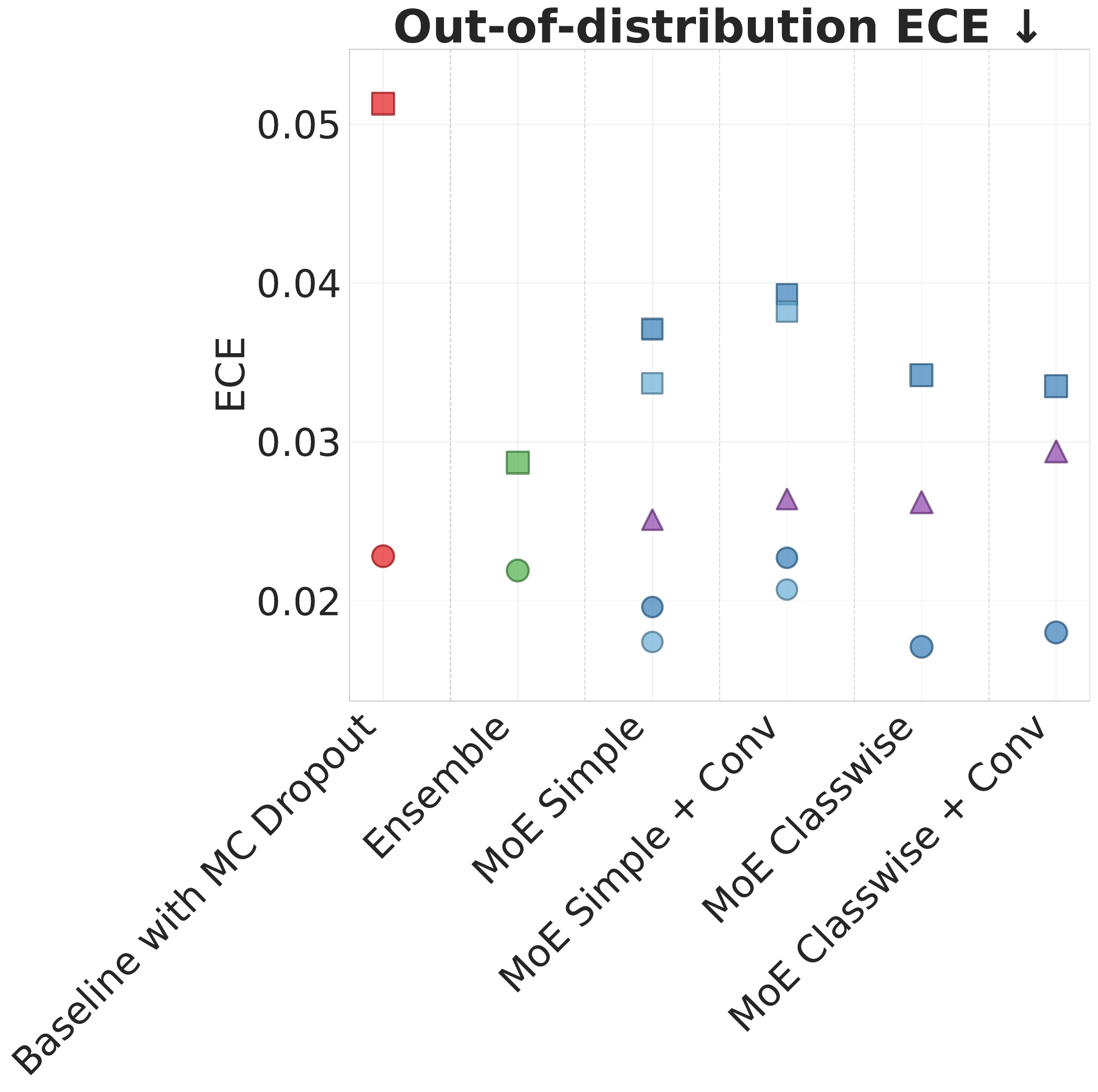}
    \end{subfigure}
    \begin{subfigure}{0.246\linewidth}
      \centering
      \includegraphics[width=\linewidth]{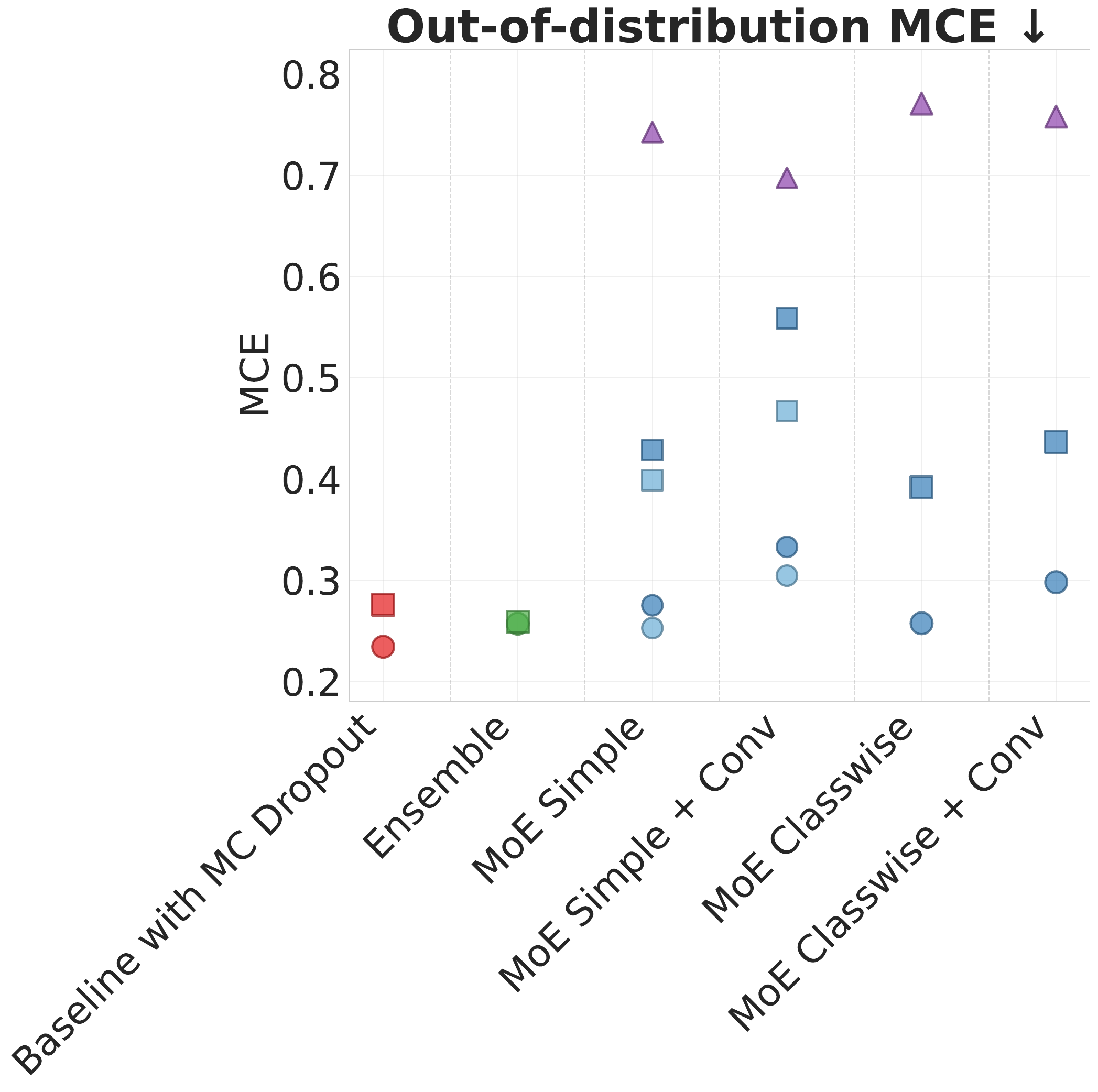}
    \end{subfigure}
    \begin{subfigure}{0.246\linewidth}
      \centering
      \includegraphics[width=\linewidth]{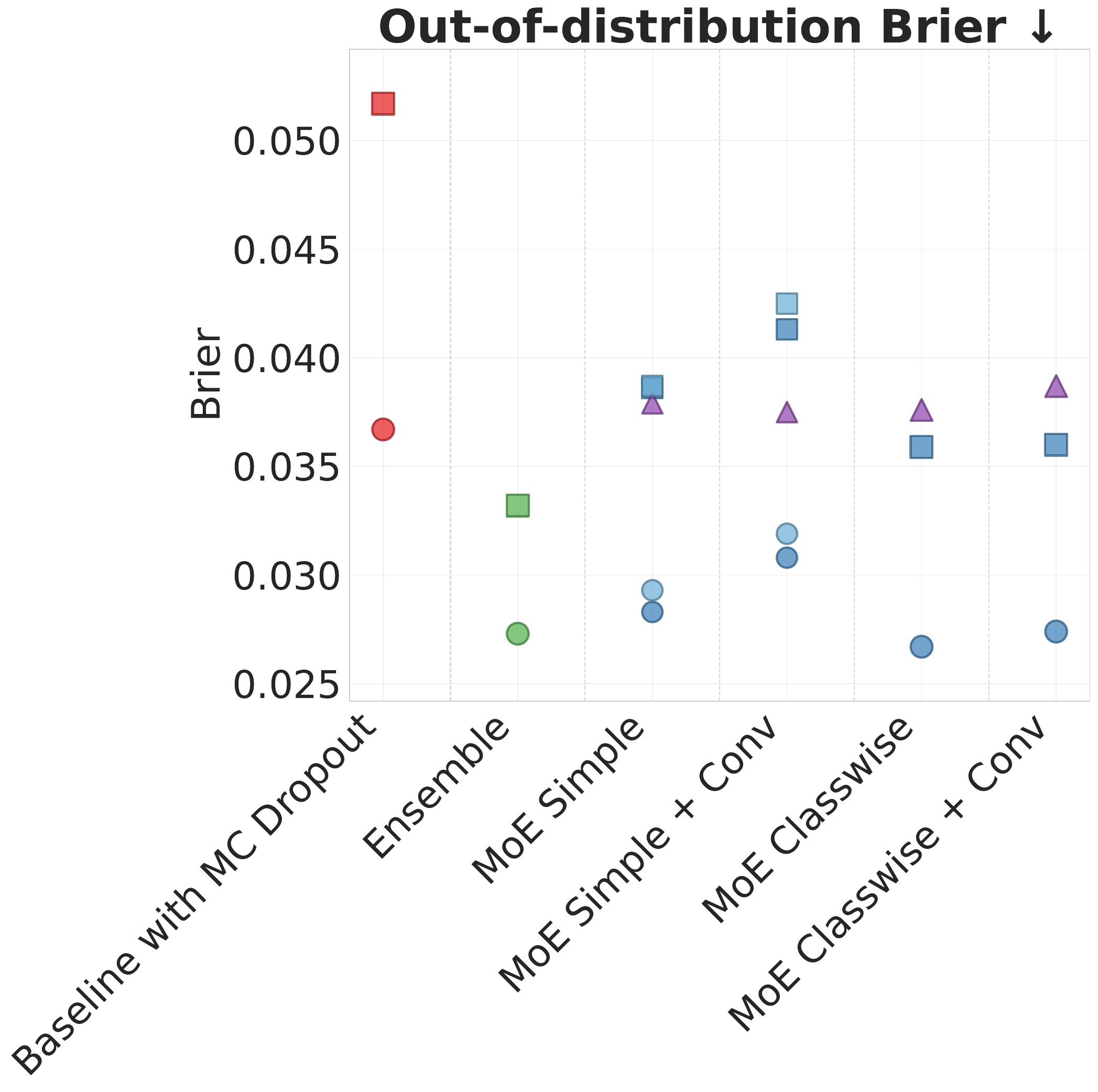}
    \end{subfigure}
    \begin{subfigure}{0.246\linewidth}
      \centering
      \includegraphics[width=\linewidth]{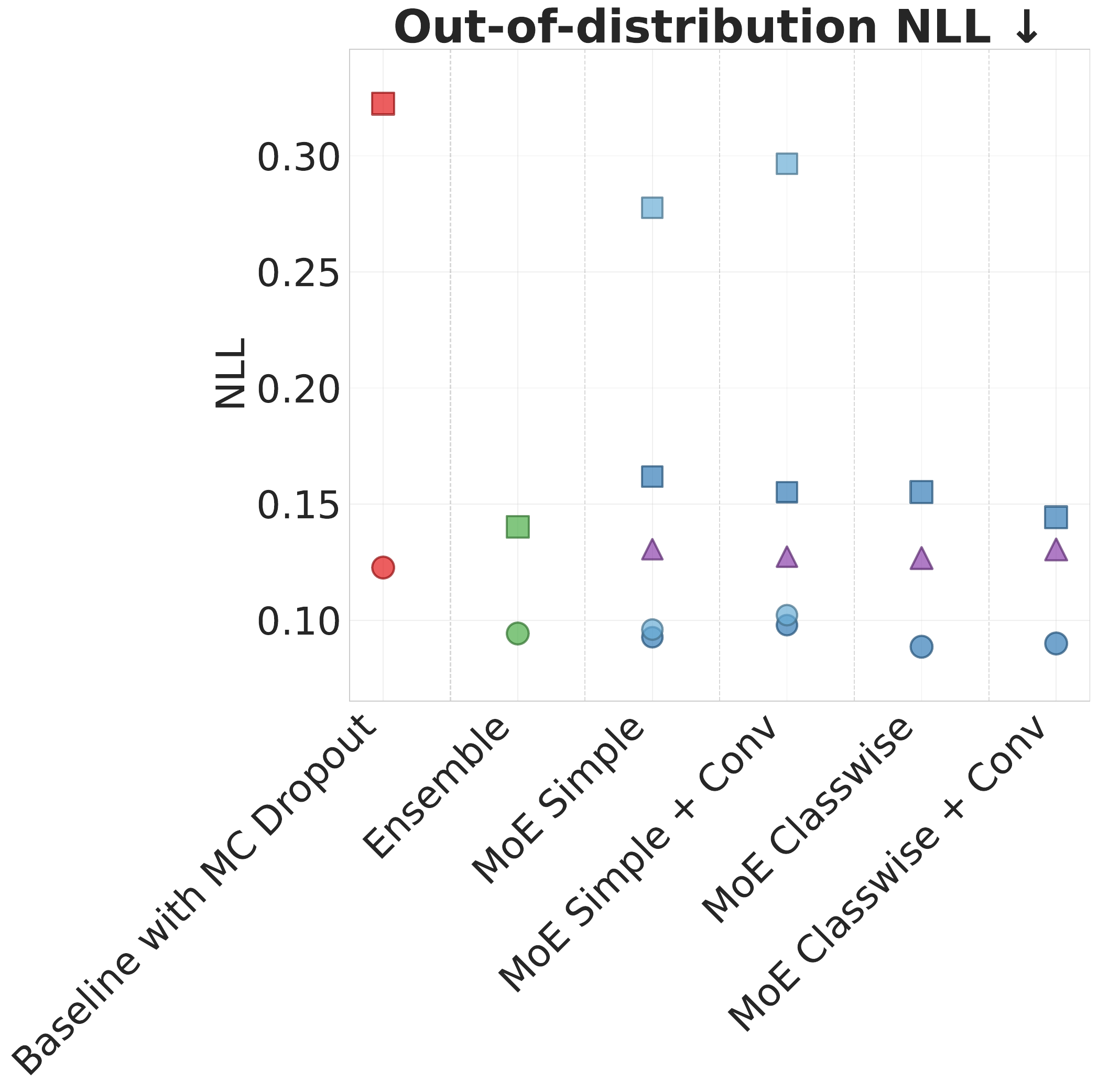}
    \end{subfigure}

    \caption{Calibration evaluation of the \textbf{predictive uncertainty} on \textbf{in-distribution (highway-urban)} and \textbf{out-of-distribution (ambiguous)} A2D2 test data. Lower values indicate better-calibrated uncertainty estimates; comparison models are shown in red.}
        \label{fig:uq_comparison}
\end{figure*}

\begin{figure*}[h]
     \centering
     \includegraphics[width=0.4\linewidth]{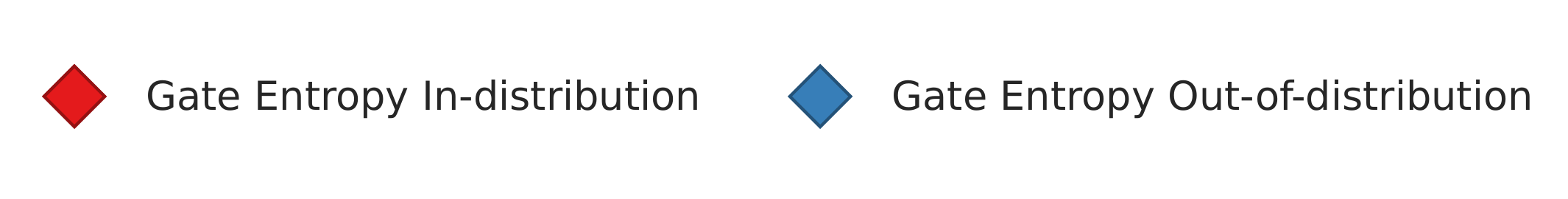}
     
     \begin{subfigure}{0.246\linewidth}
      \centering
      \includegraphics[width=\linewidth]{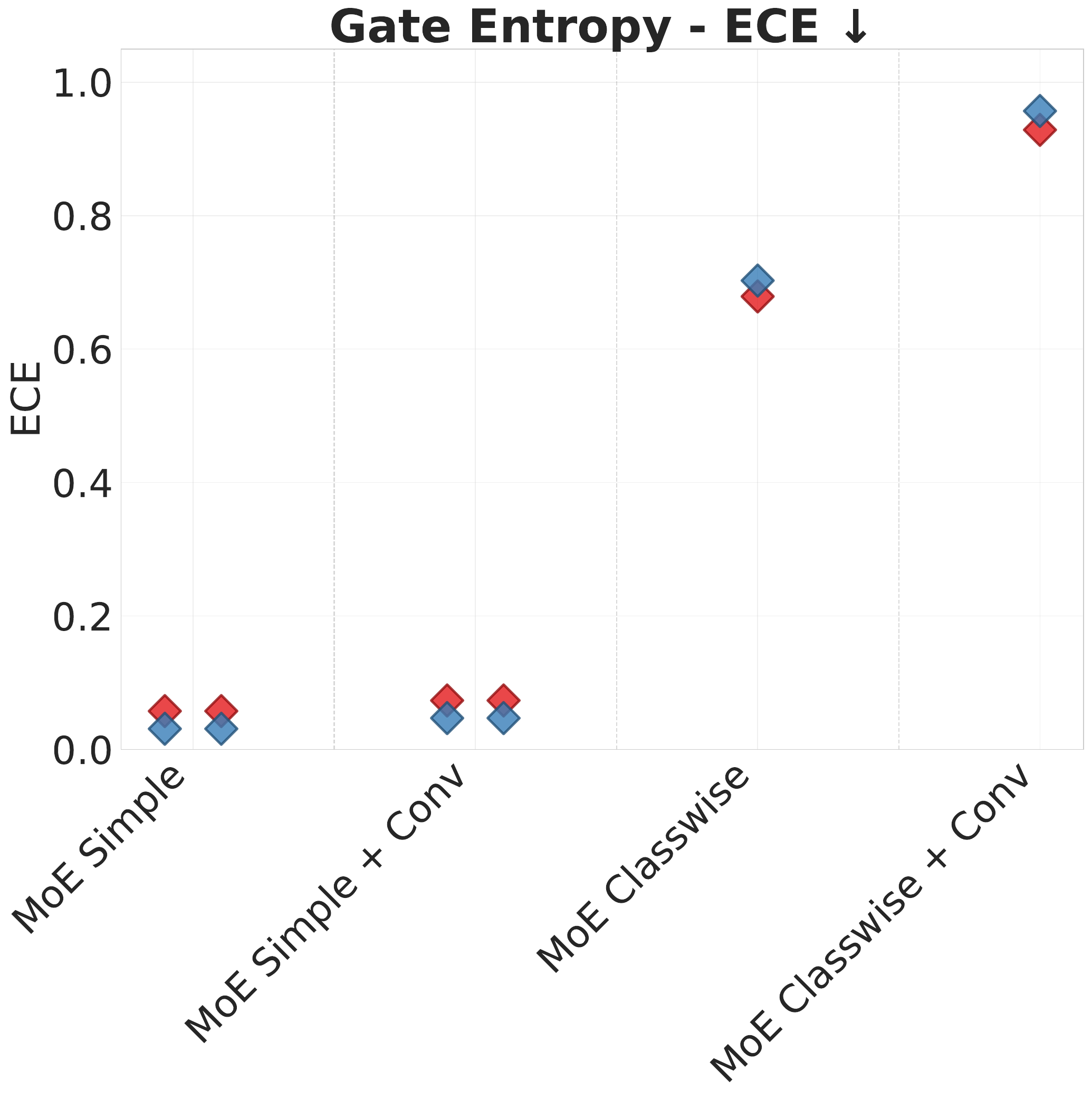}
    \end{subfigure}
    \begin{subfigure}{0.246\linewidth}
      \centering
      \includegraphics[width=\linewidth]{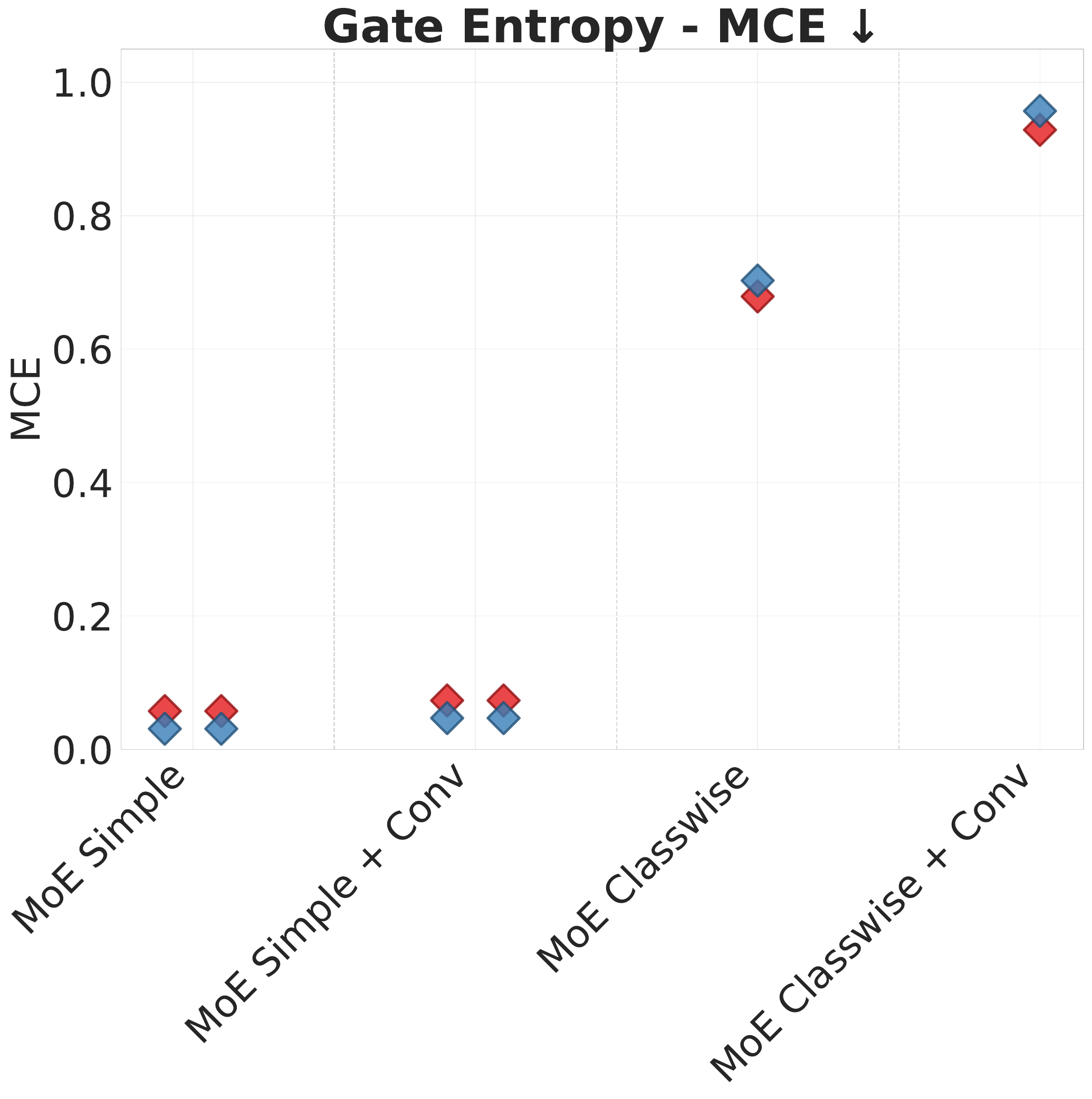}
    \end{subfigure}
    \begin{subfigure}{0.246\linewidth}
      \centering
      \includegraphics[width=\linewidth]{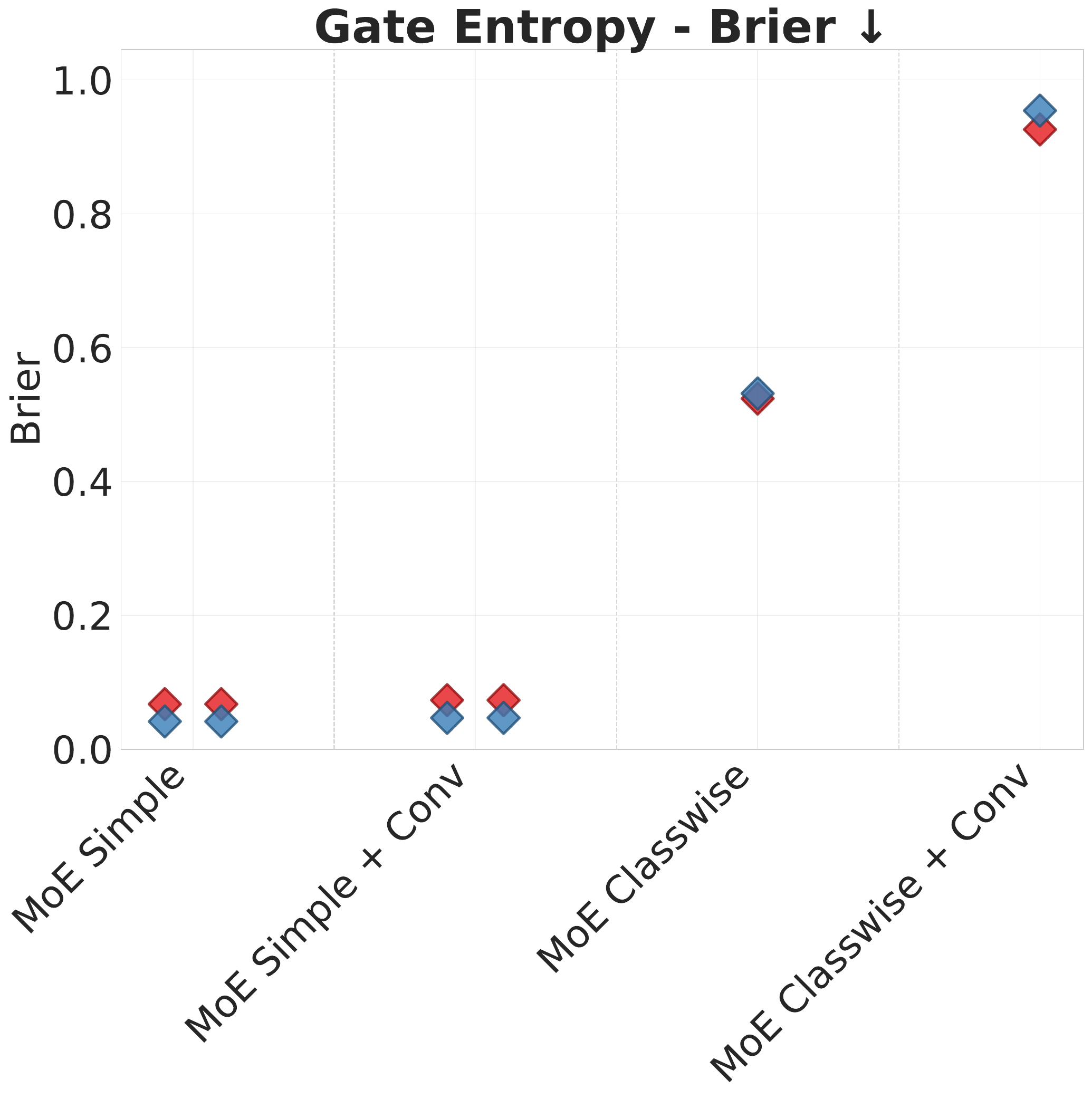}
    \end{subfigure}
    \begin{subfigure}{0.246\linewidth}
      \centering
      \includegraphics[width=\linewidth]{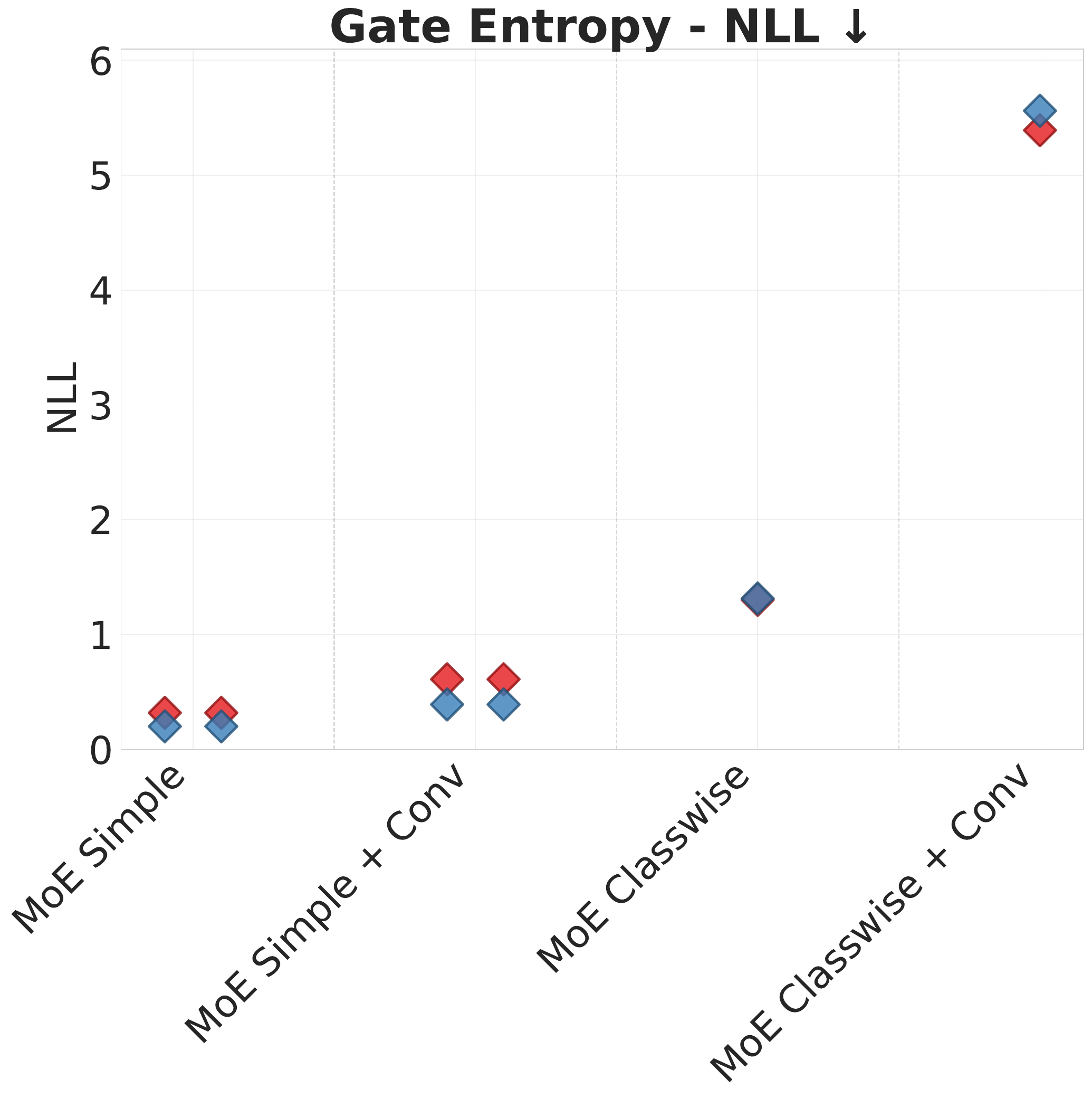}
    \end{subfigure}

    \caption{Calibration evaluation of the \textbf{routing uncertainty} on \textbf{in-distribution (highway-urban)} and \textbf{out-of-distribution (ambiguous)} A2D2 test data. Lower values indicate better-calibrated uncertainty estimates.}
        \label{fig:gate-entropy}
\end{figure*}

\begin{figure}[t]
    \centering
    \includegraphics[width=0.7\linewidth]{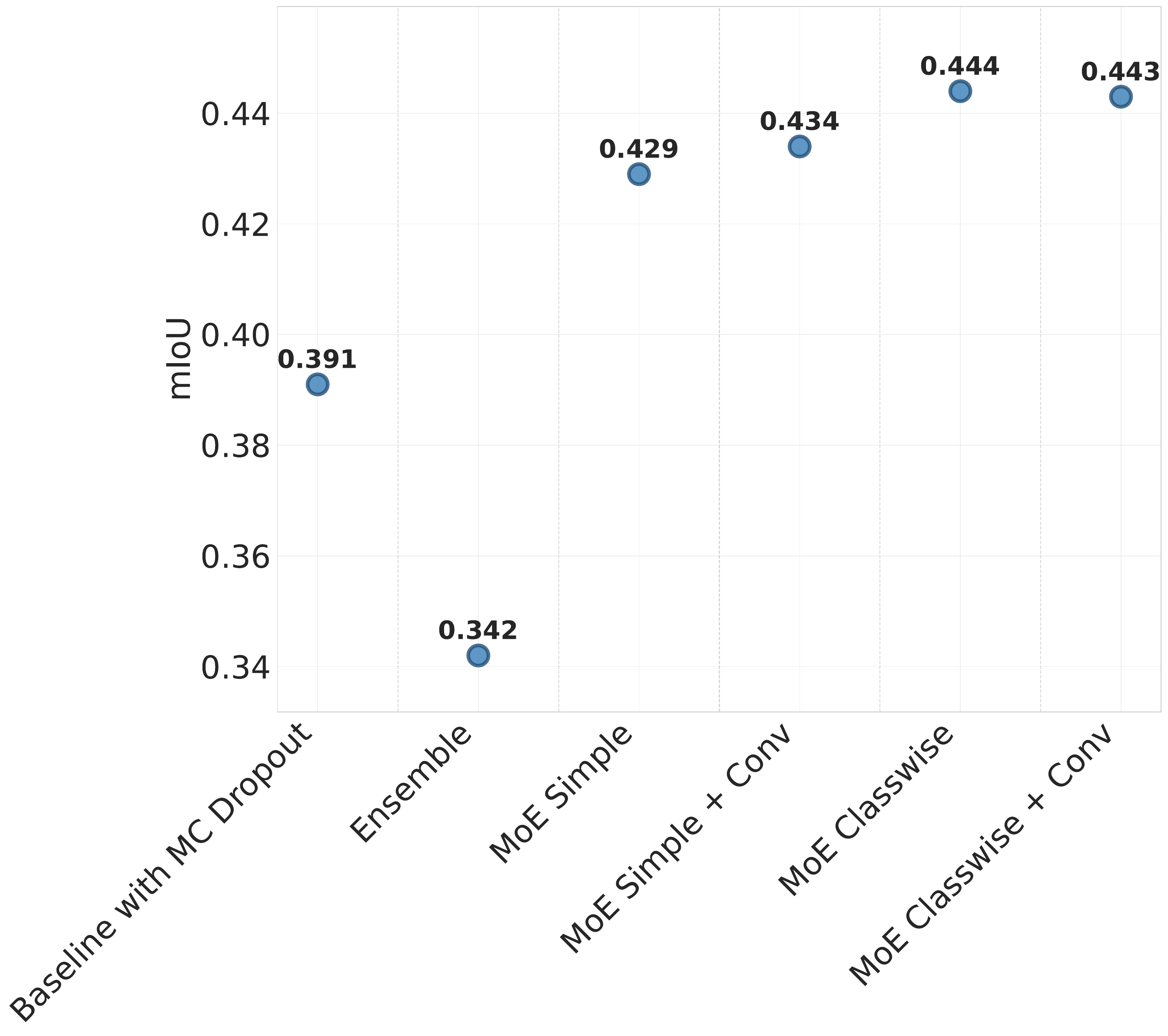}
    \caption{Semantic segmentation performance on A2D2 test data. Higher values indicate better semantic segmentation accuracy.}
    \label{fig:miou}
\end{figure}

To implement MC dropout for the \texttt{DeepLabv3+} model, the existing dropout layers are activated during testing, and further dropout layers are added to the backbone's output features within each branch of the ASPP module, including its convolutional and pooling components, immediately before the final classification layer. We evaluated different dropout rates (see Figure~\ref{fig:dropout-rate}), indicating better calibration for higher values, but at the cost of lower model performance. For the experiments, a dropout rate of 0.1 is thus selected, which matches the original \texttt{DeepLabv3+} dropout rate.
 
\begin{table*}[t]
\centering
\caption{Qualitative results for the uncertainty quantification methods across three driving environments.}
\label{tab:uq_examples}
\resizebox{\linewidth}{!}{%
  \begin{tabular}{r c c c c c c c c c c}
      & \textbf{Input} & \textbf{Ground truth} 
      & \multicolumn{2}{c}{\textbf{Baseline (MC Dropout)}} 
      & \multicolumn{2}{c}{\textbf{Ensemble}}
      & \multicolumn{3}{c}{\textbf{MoE, classwise gate}}
      &  \\ 
      
      &&& PE & MI & PE & MI & PE & MI & EV
      & \multirow[c]{4}{*}[2pt]{%
          \adjustbox{valign=c}{%
            \includegraphics[height=0.37\textwidth]{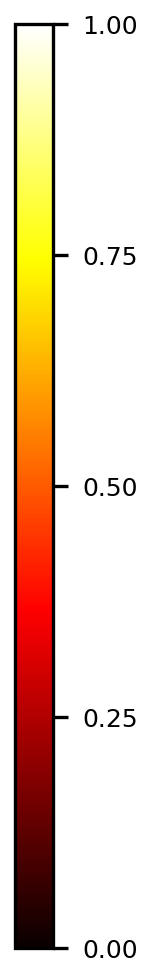}
          }
        } \\
      \\[4pt]
    \textbf{Highway} &
      \includegraphics[width=0.1\textwidth]{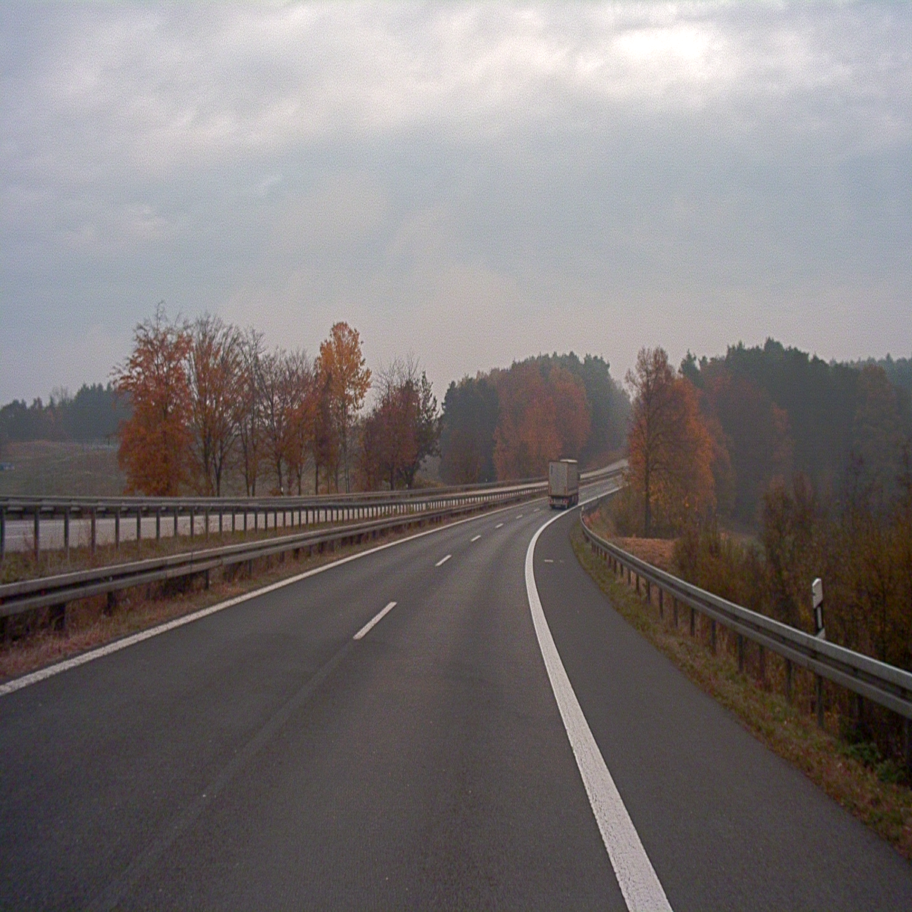} &
      \includegraphics[width=0.1\textwidth]{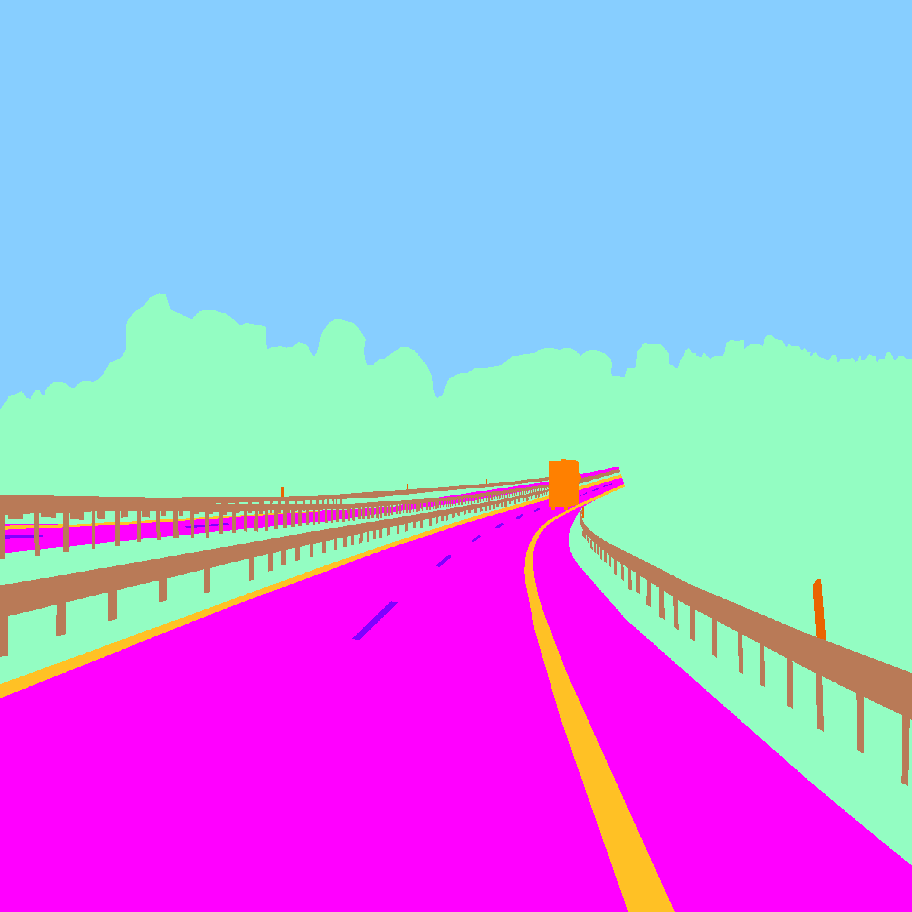} &
      \includegraphics[width=0.1\textwidth]{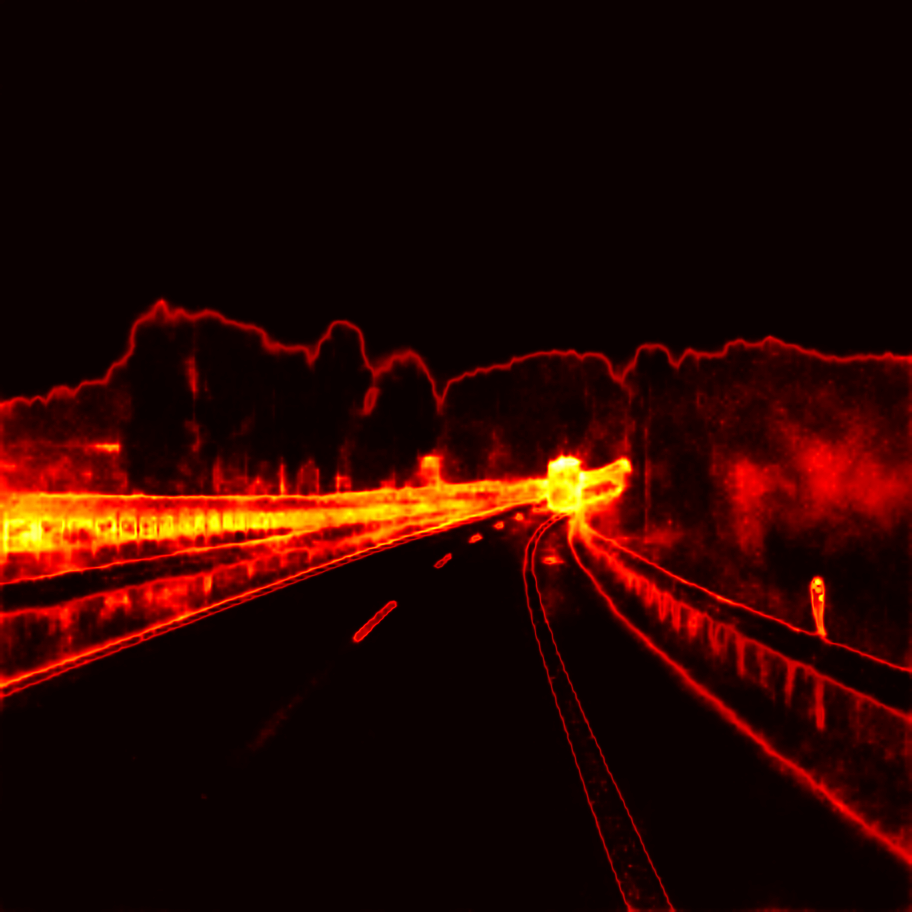} &
      \includegraphics[width=0.1\textwidth]{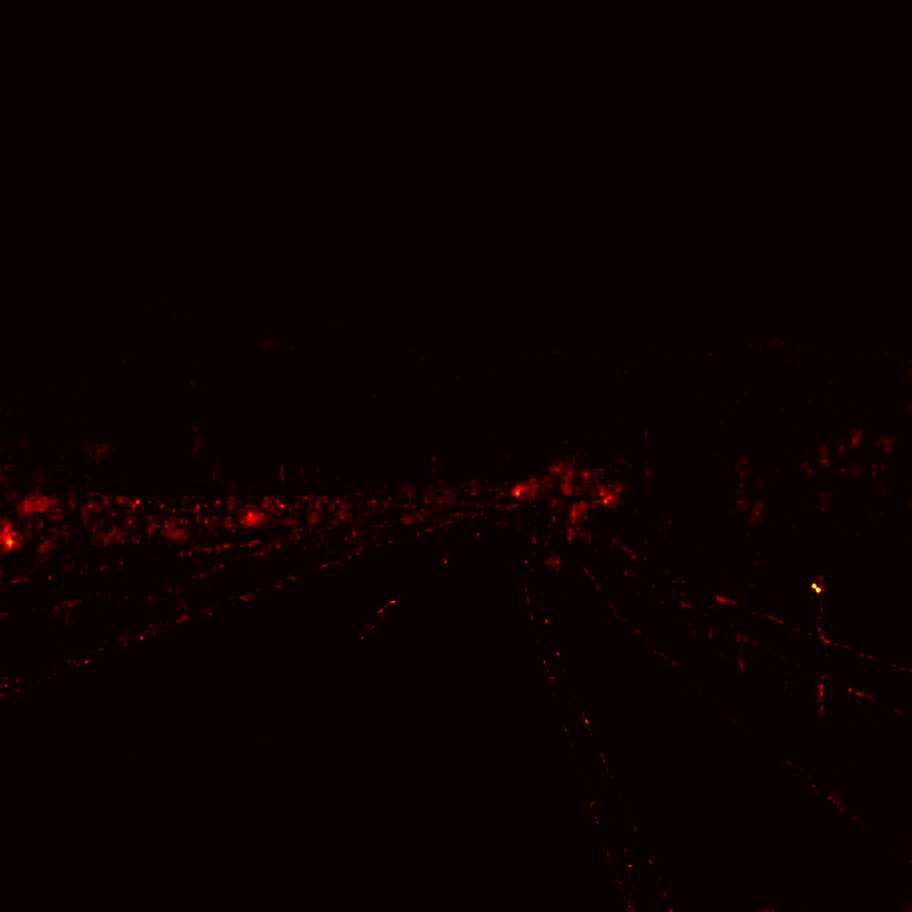} &
      \includegraphics[width=0.1\textwidth]{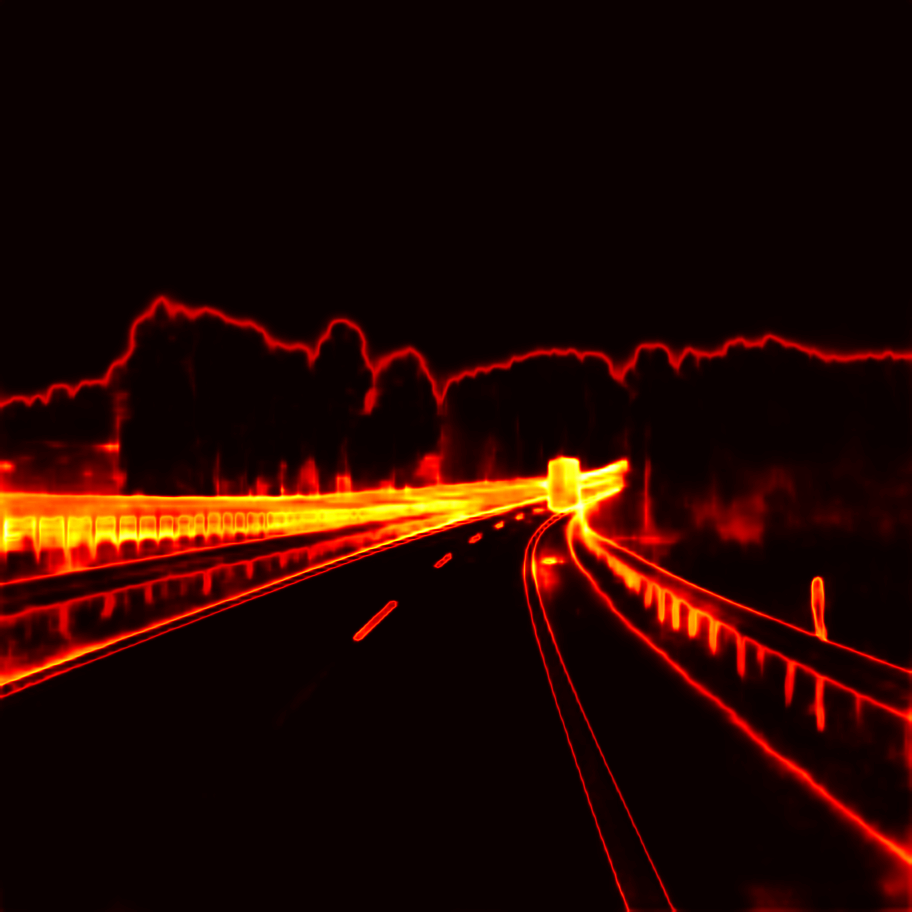} &
      \includegraphics[width=0.1\textwidth]{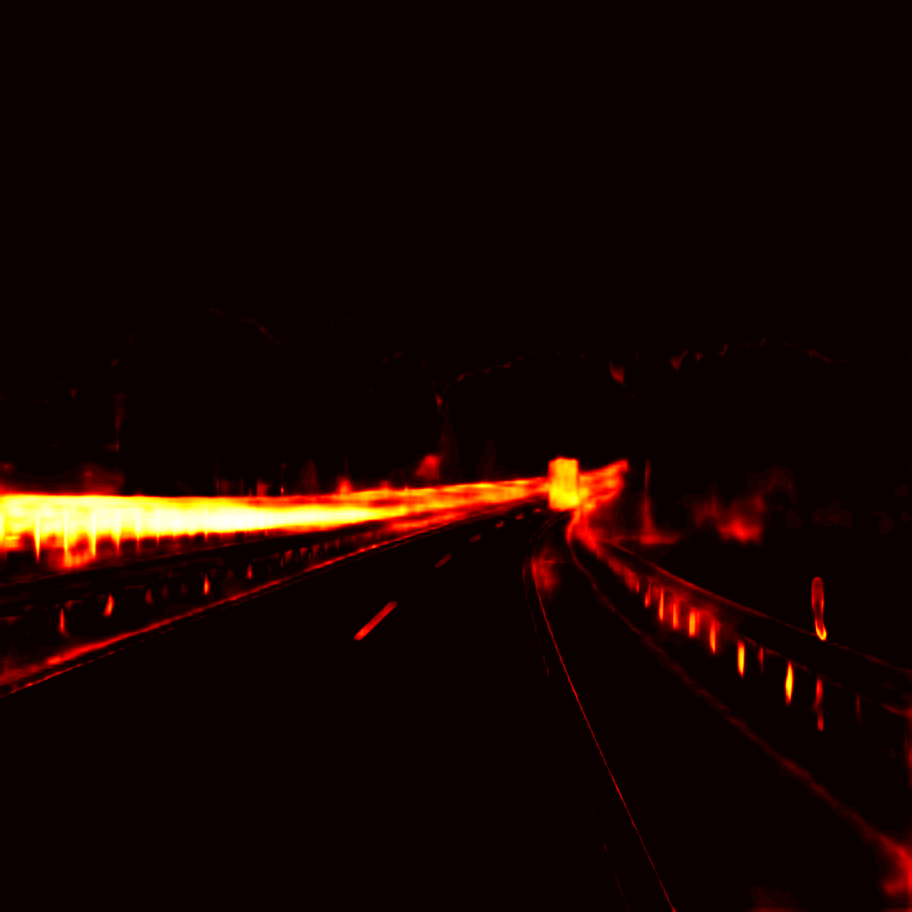} &
      \includegraphics[width=0.1\textwidth]{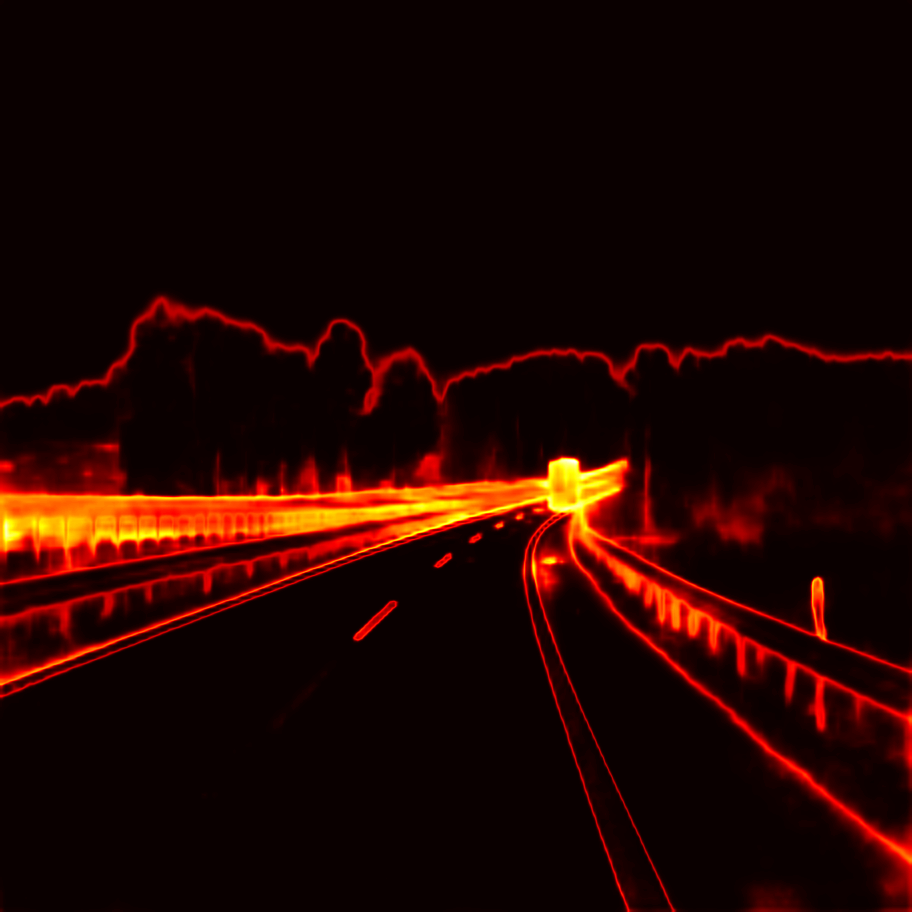} &
      \includegraphics[width=0.1\textwidth]{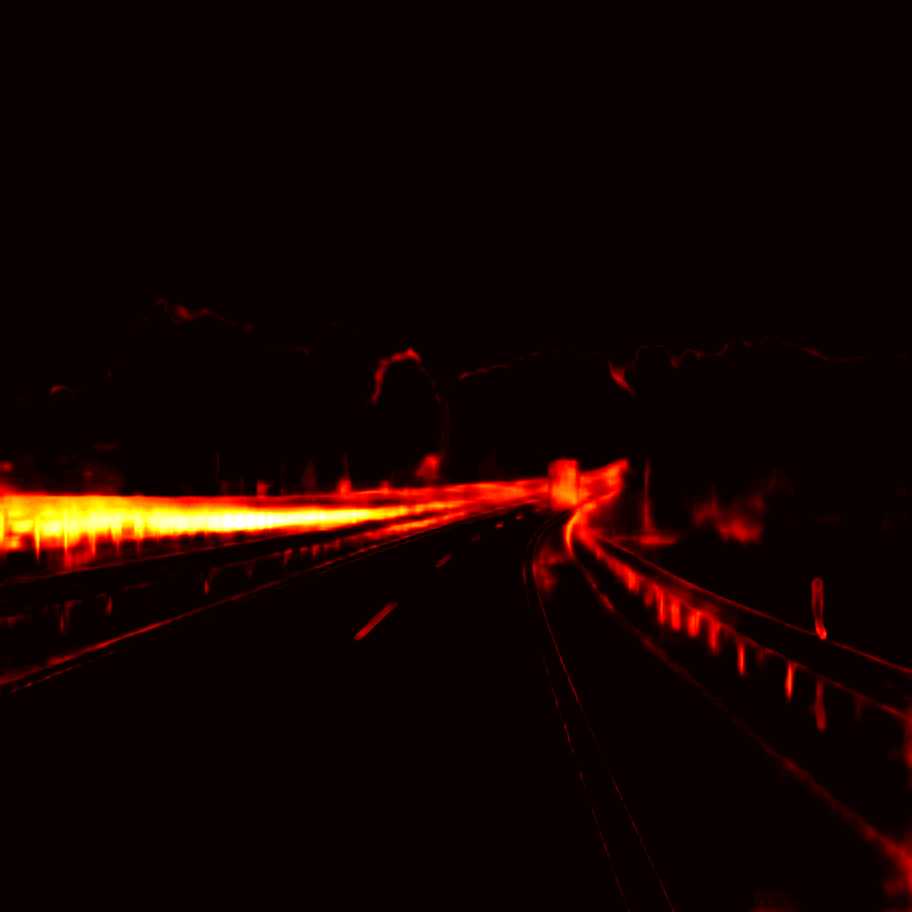} &
      \includegraphics[width=0.1\textwidth]{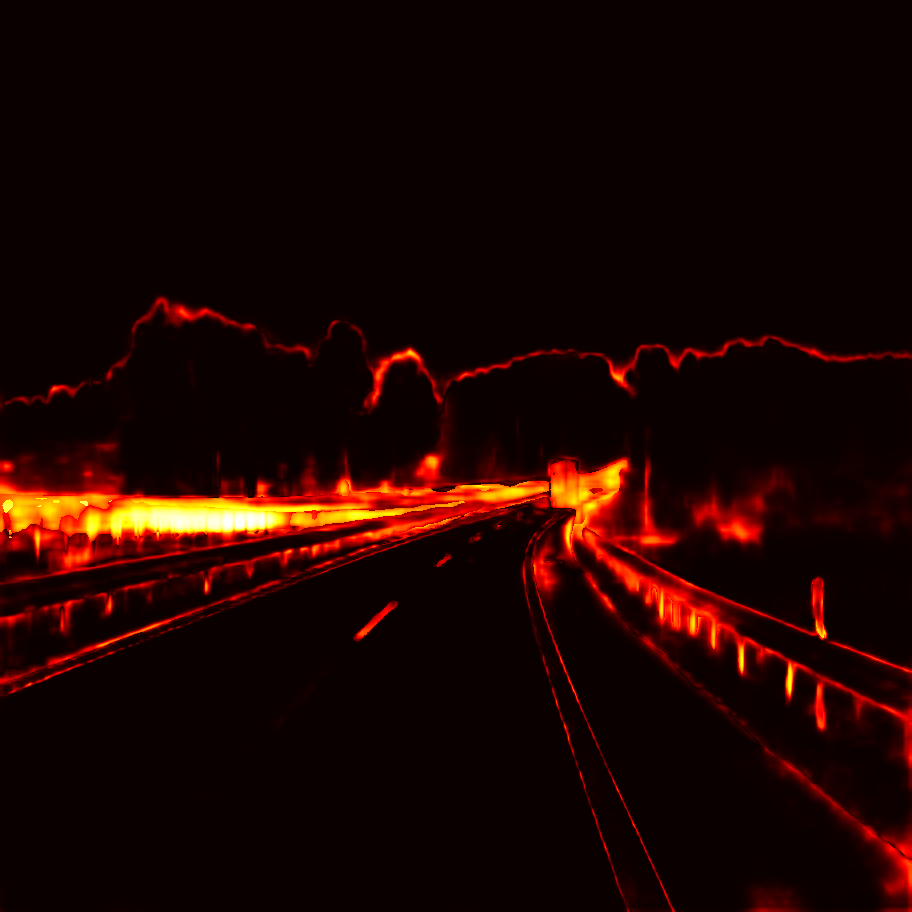} 
      &  \\

    \textbf{Urban} &
      \includegraphics[width=0.1\textwidth]{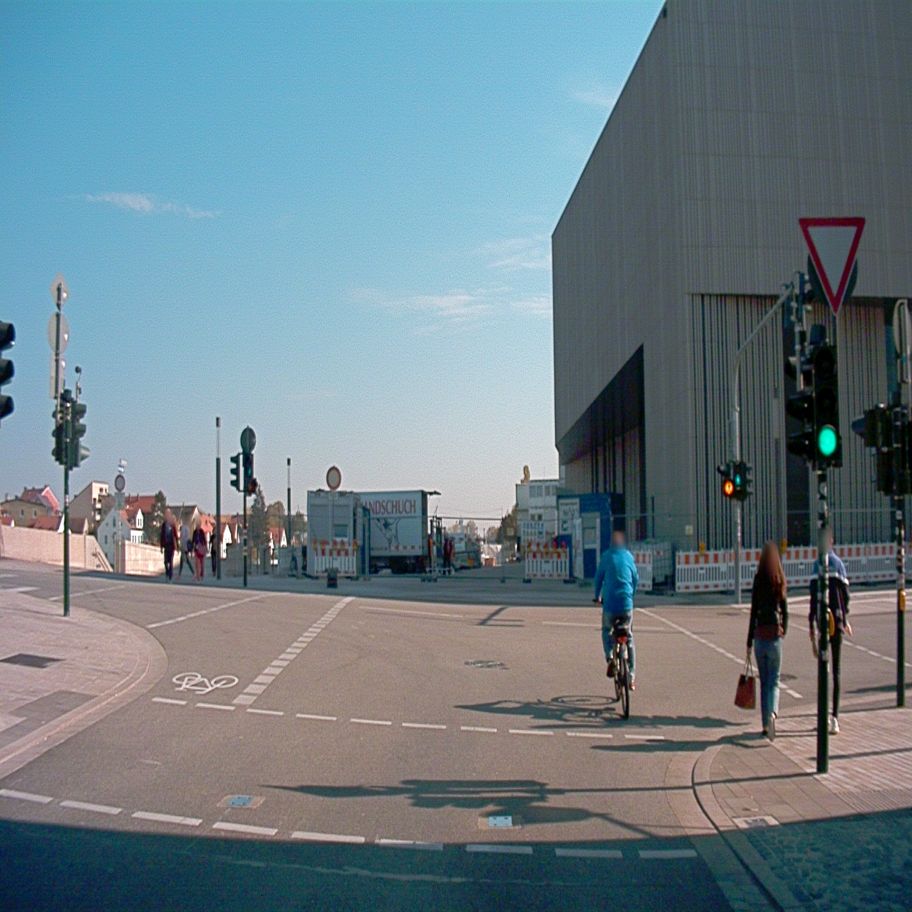} &
      \includegraphics[width=0.1\textwidth]{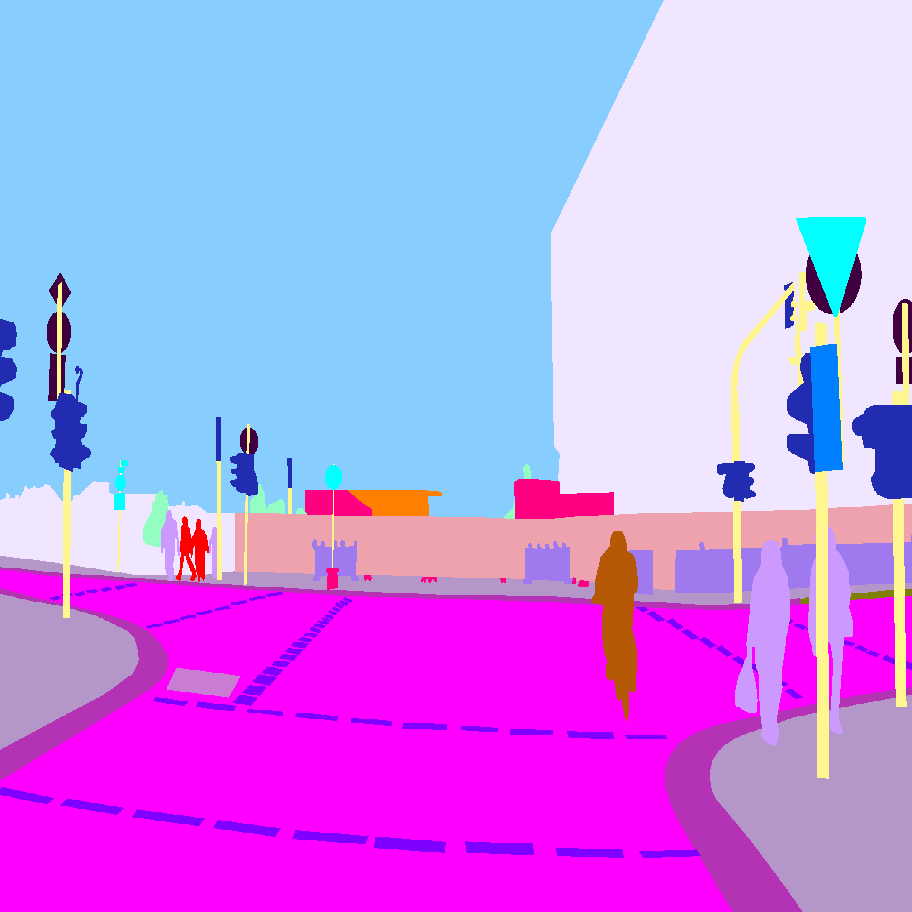} &
      \includegraphics[width=0.1\textwidth]{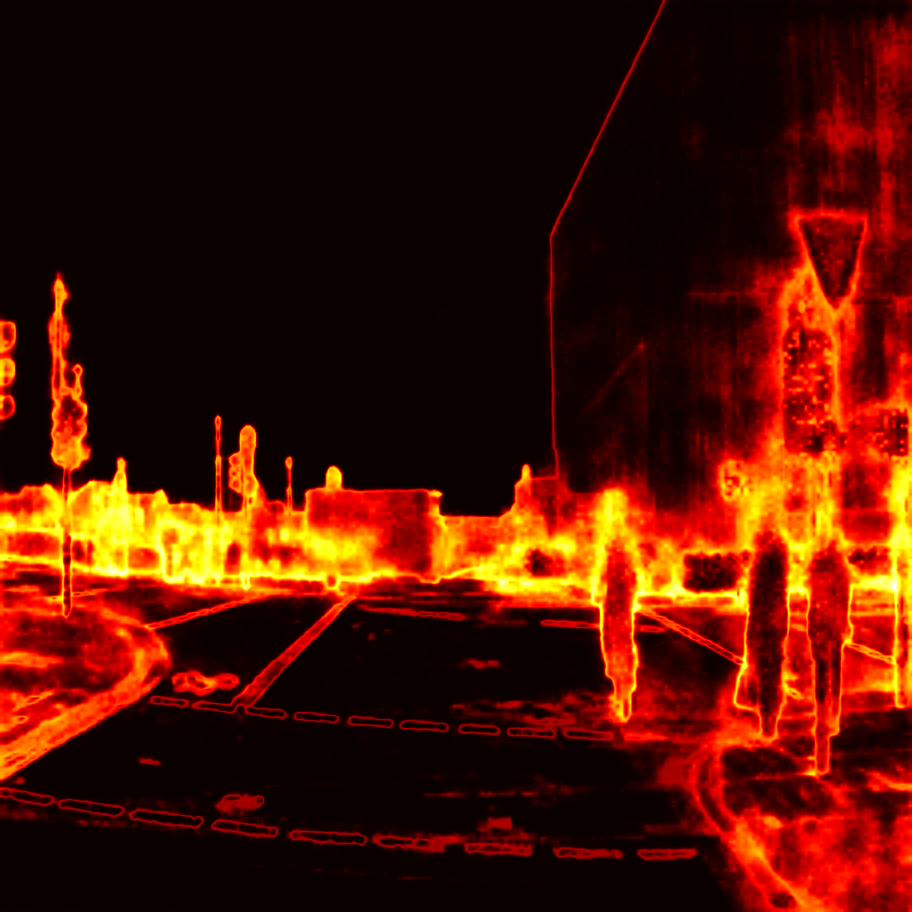} &
      \includegraphics[width=0.1\textwidth]{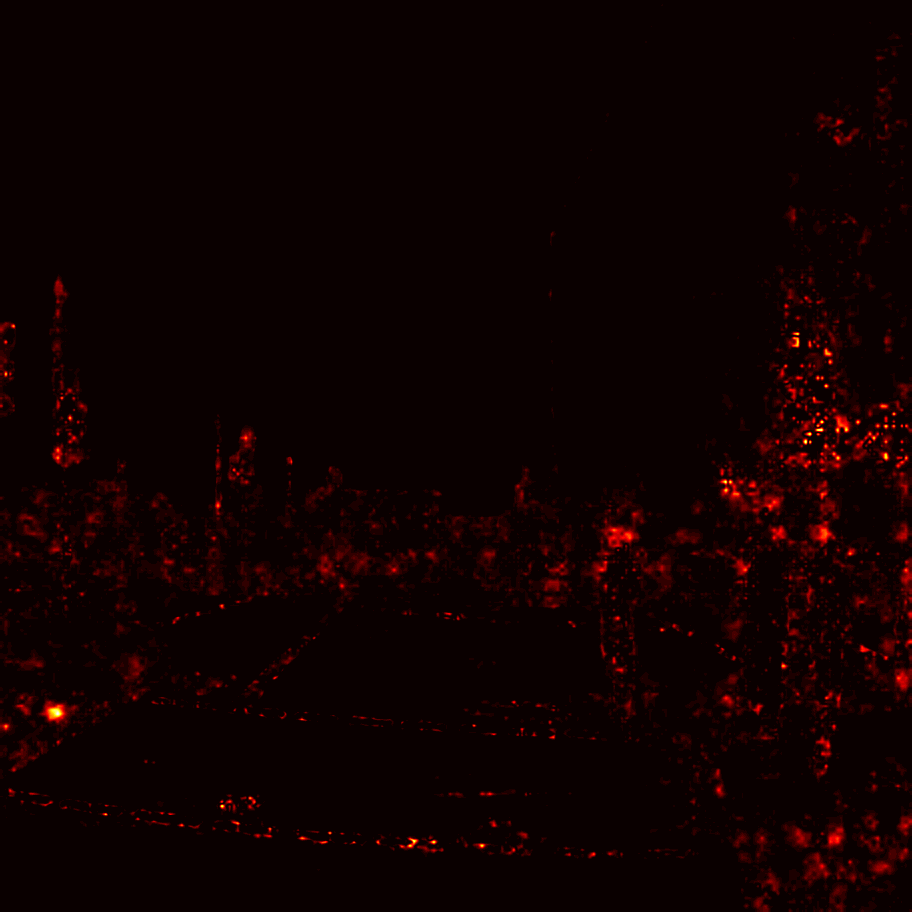} &
      \includegraphics[width=0.1\textwidth]{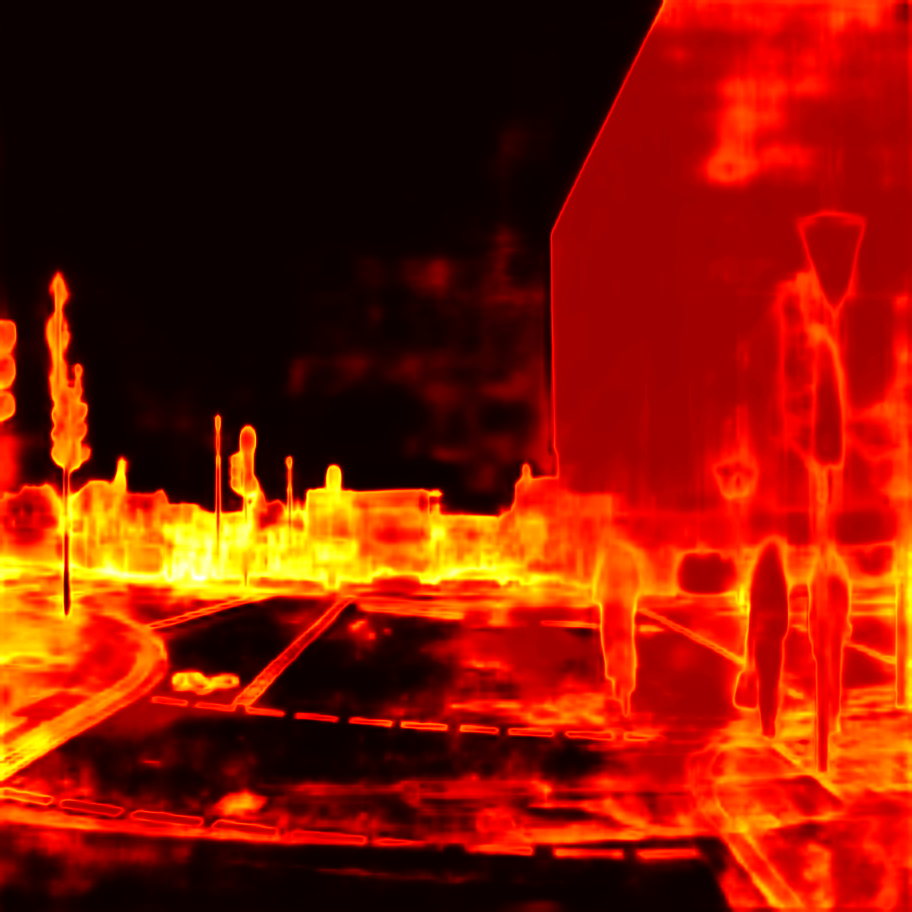} &
      \includegraphics[width=0.1\textwidth]{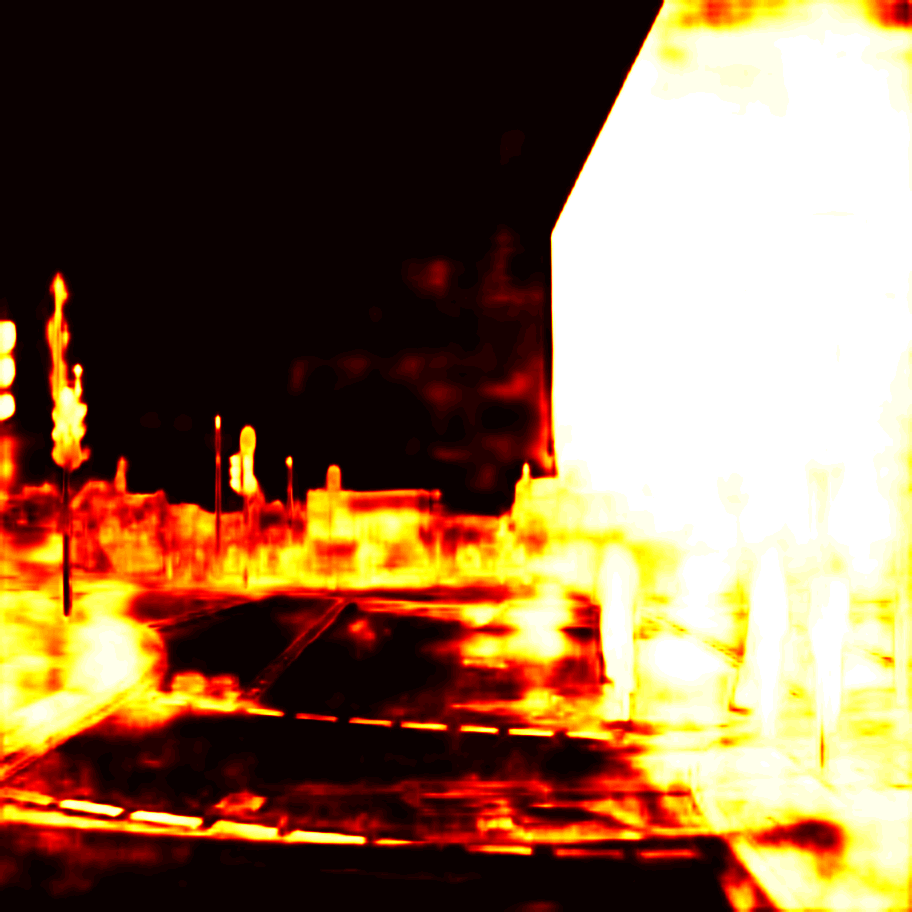} &
      \includegraphics[width=0.1\textwidth]{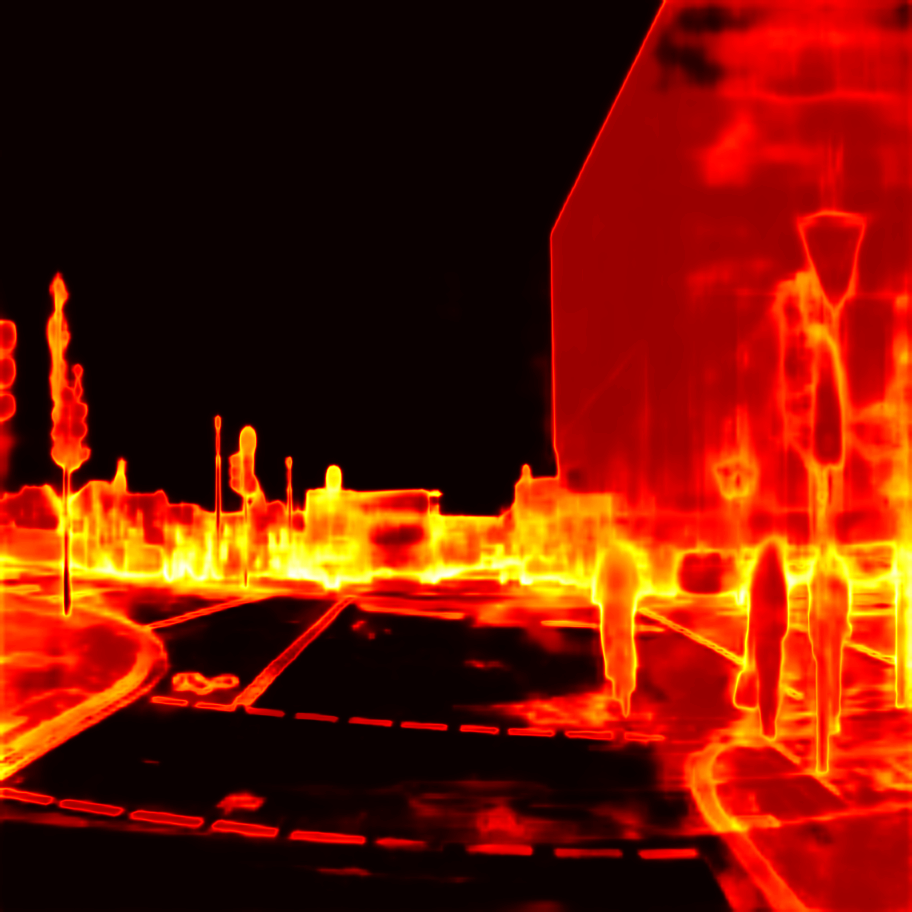} &
      \includegraphics[width=0.1\textwidth]{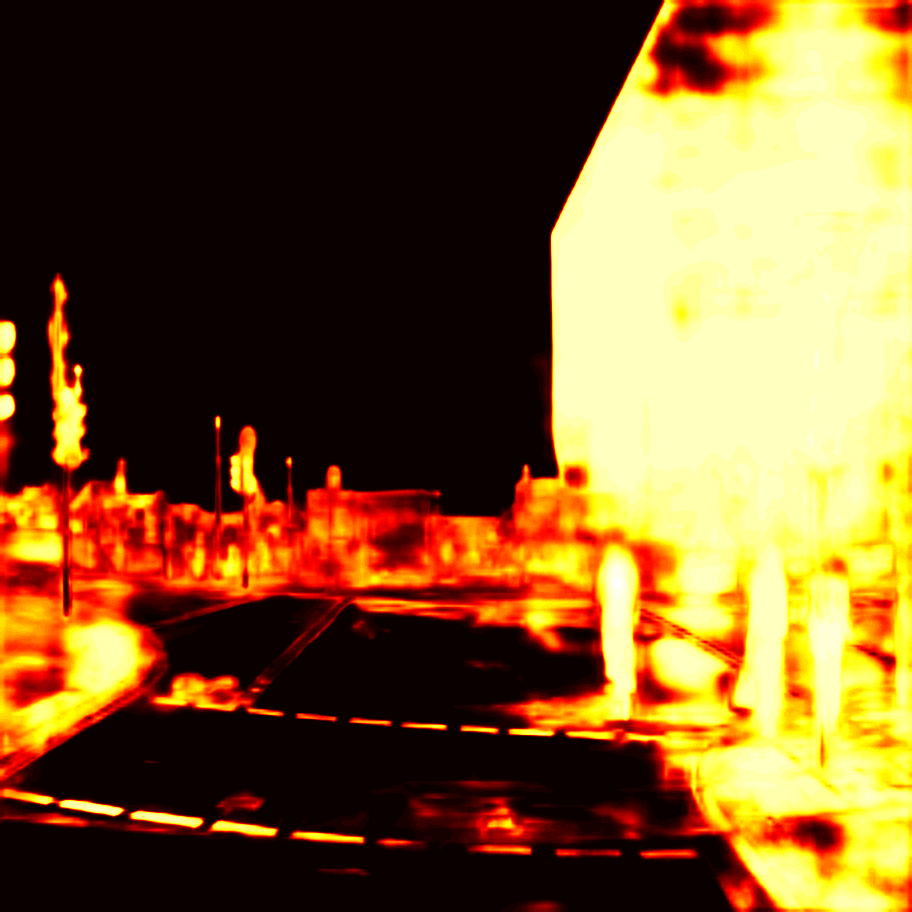} &
      \includegraphics[width=0.1\textwidth]{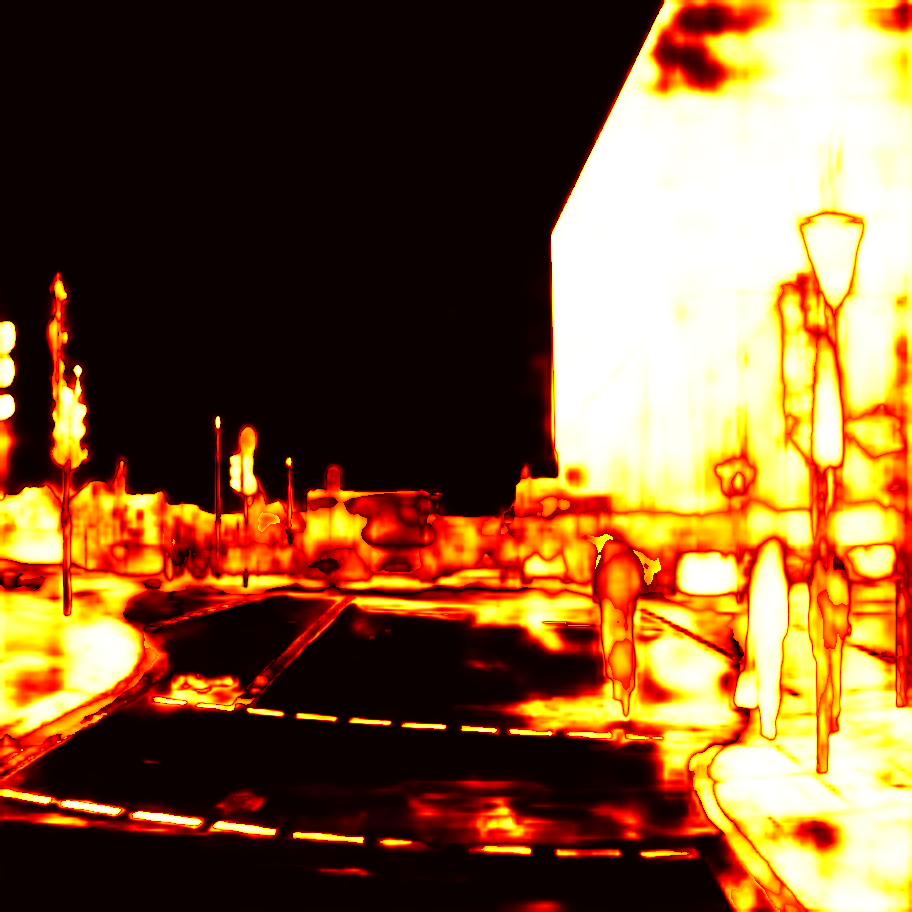} 
      &  \\

    \textbf{Ambiguous} &
      \includegraphics[width=0.1\textwidth]{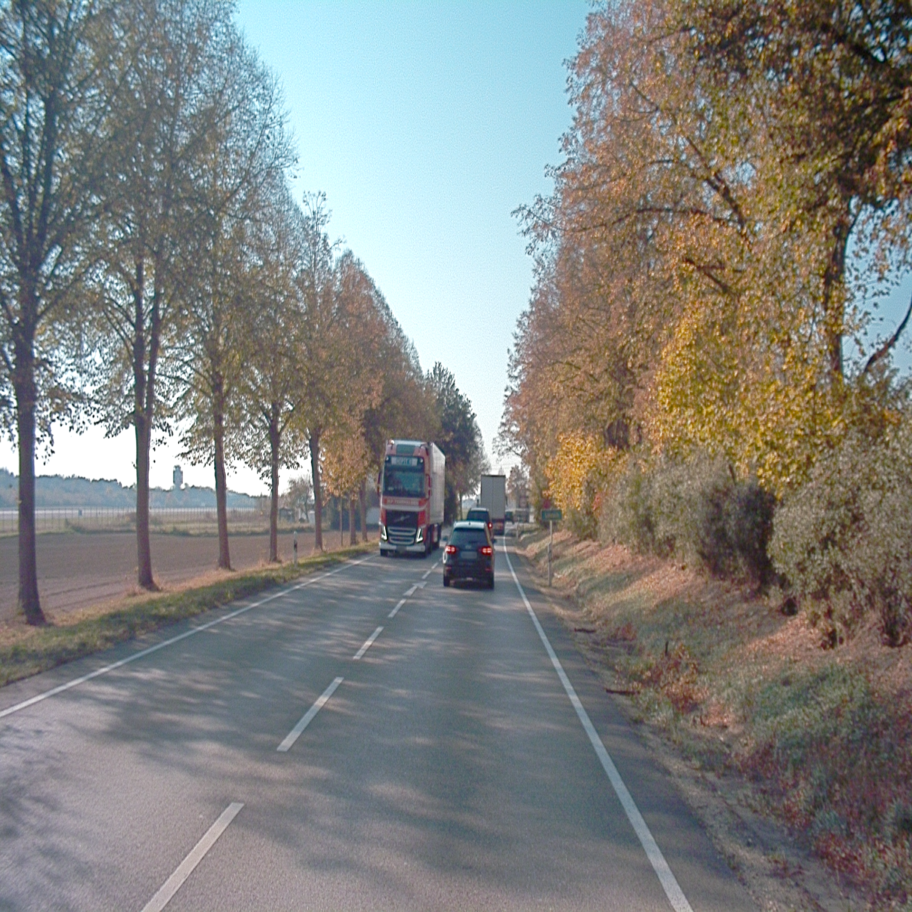} &
      \includegraphics[width=0.1\textwidth]{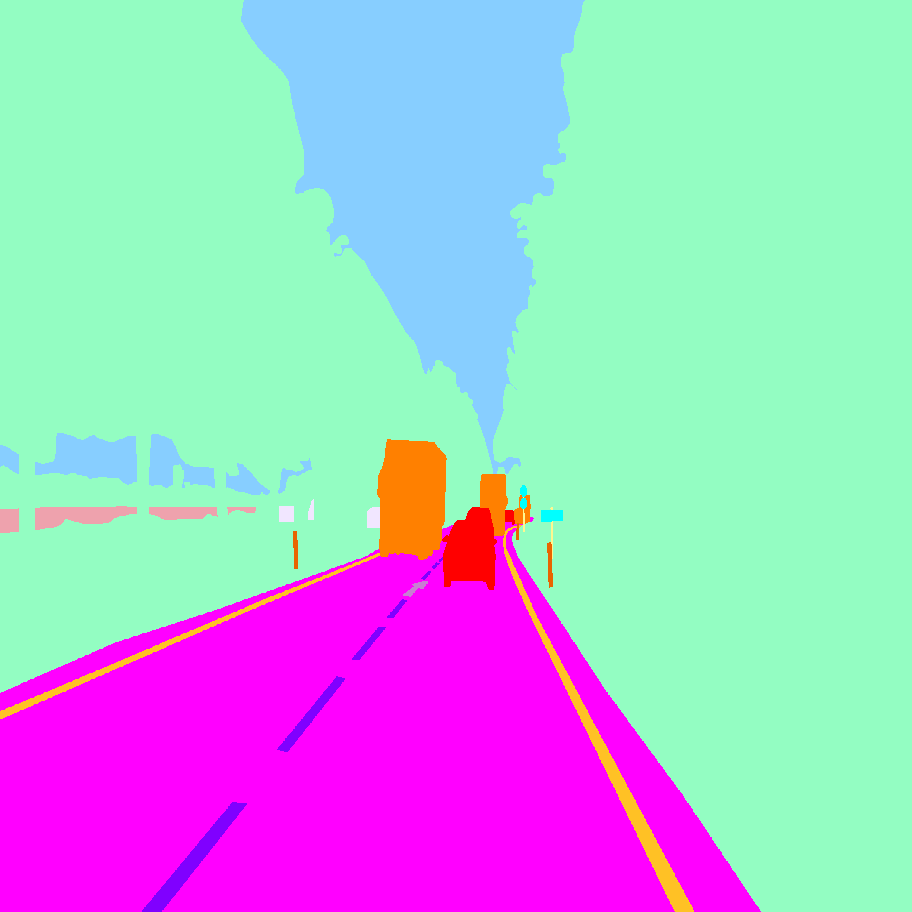} &
      \includegraphics[width=0.1\textwidth]{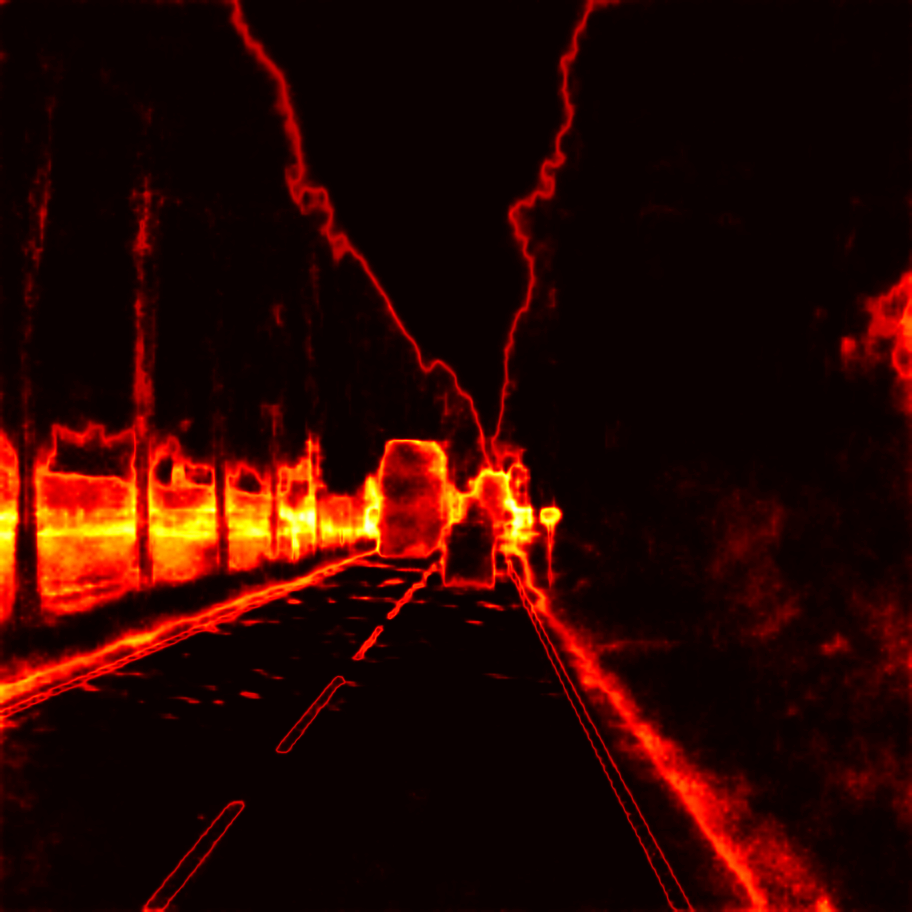} &
      \includegraphics[width=0.1\textwidth]{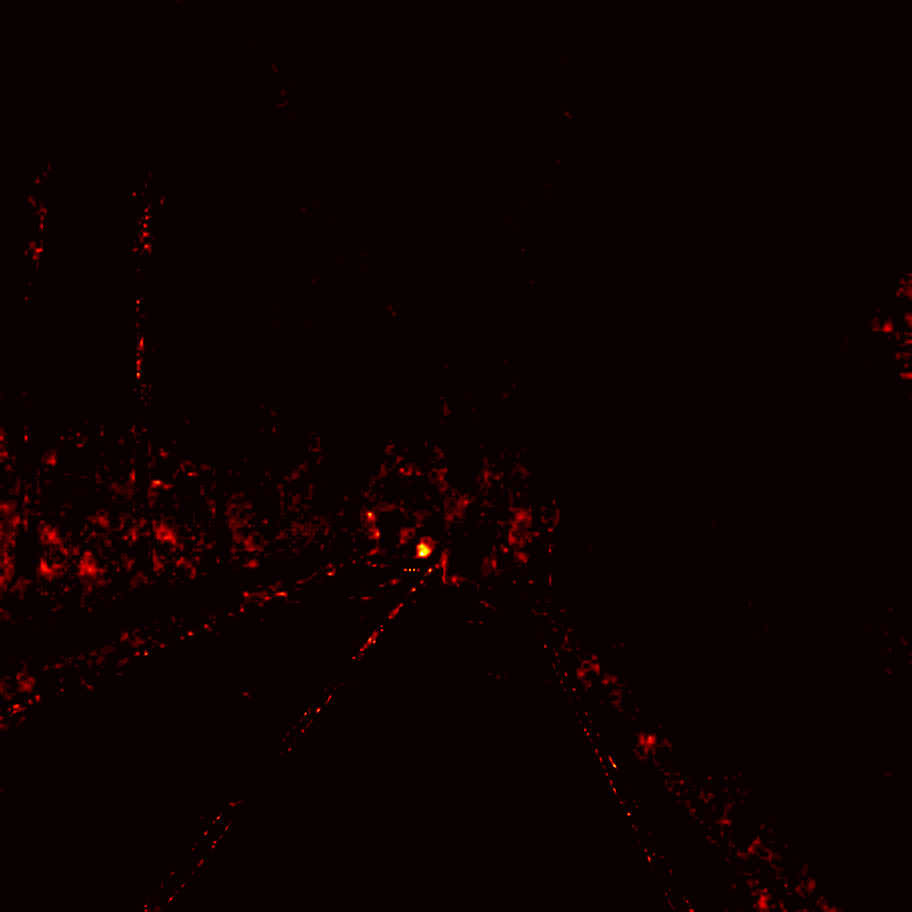} &
      \includegraphics[width=0.1\textwidth]{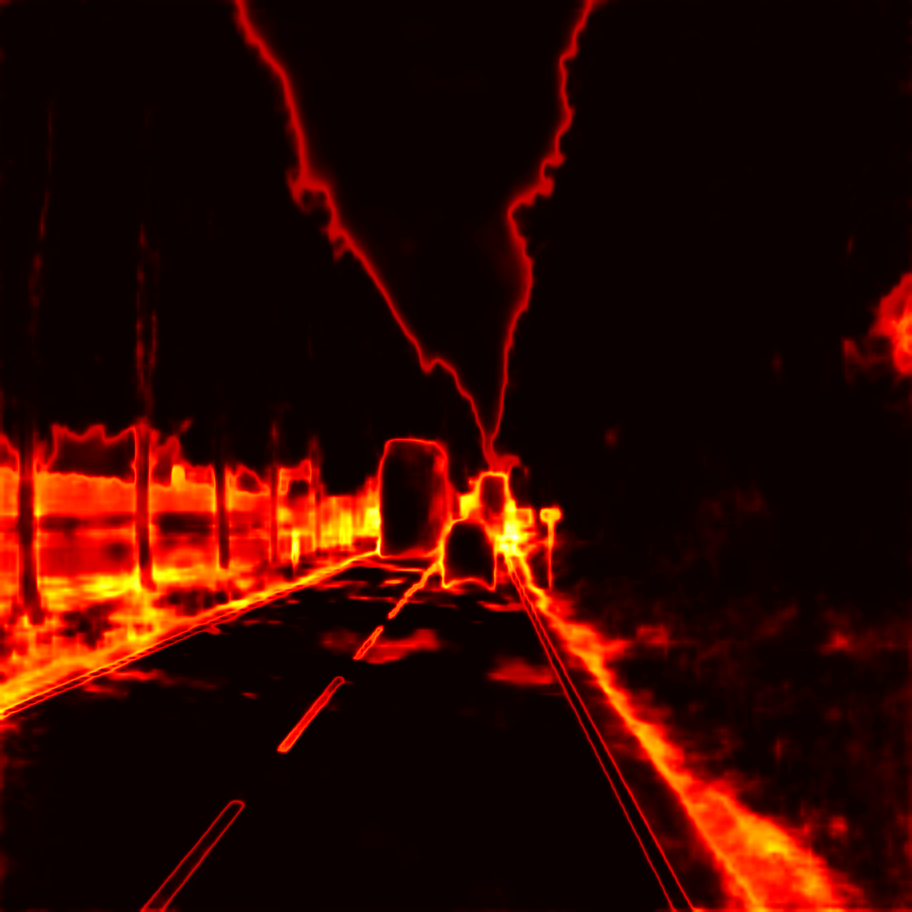} &
      \includegraphics[width=0.1\textwidth]{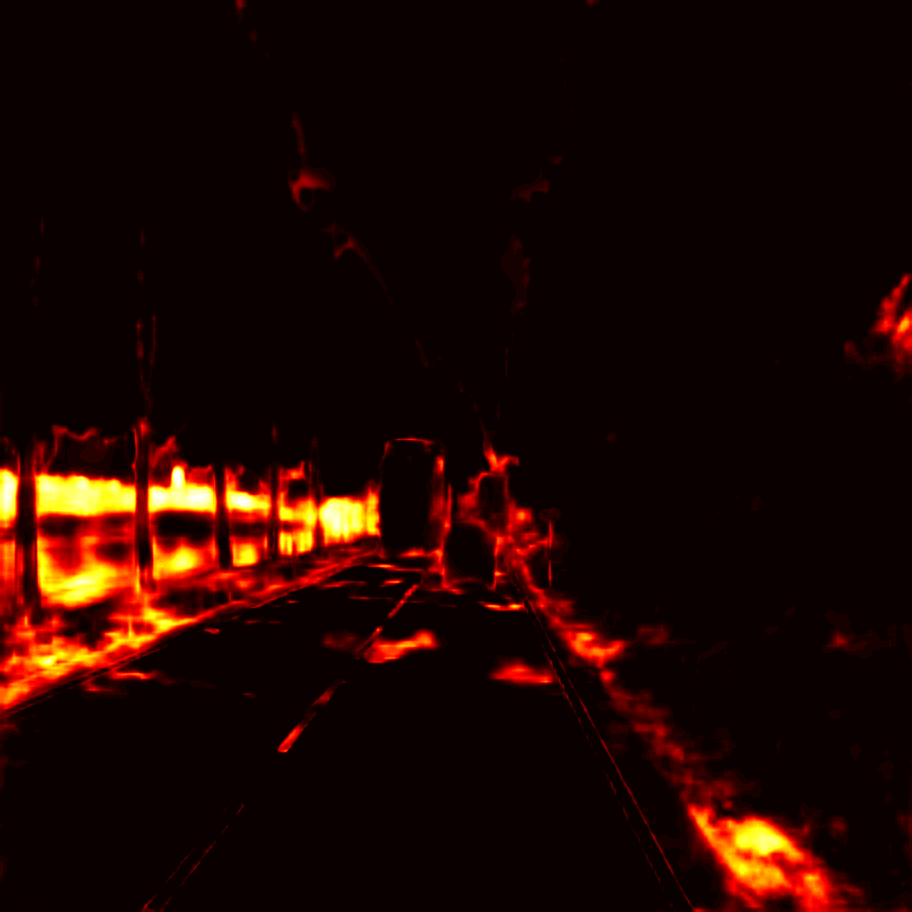} &
      \includegraphics[width=0.1\textwidth]{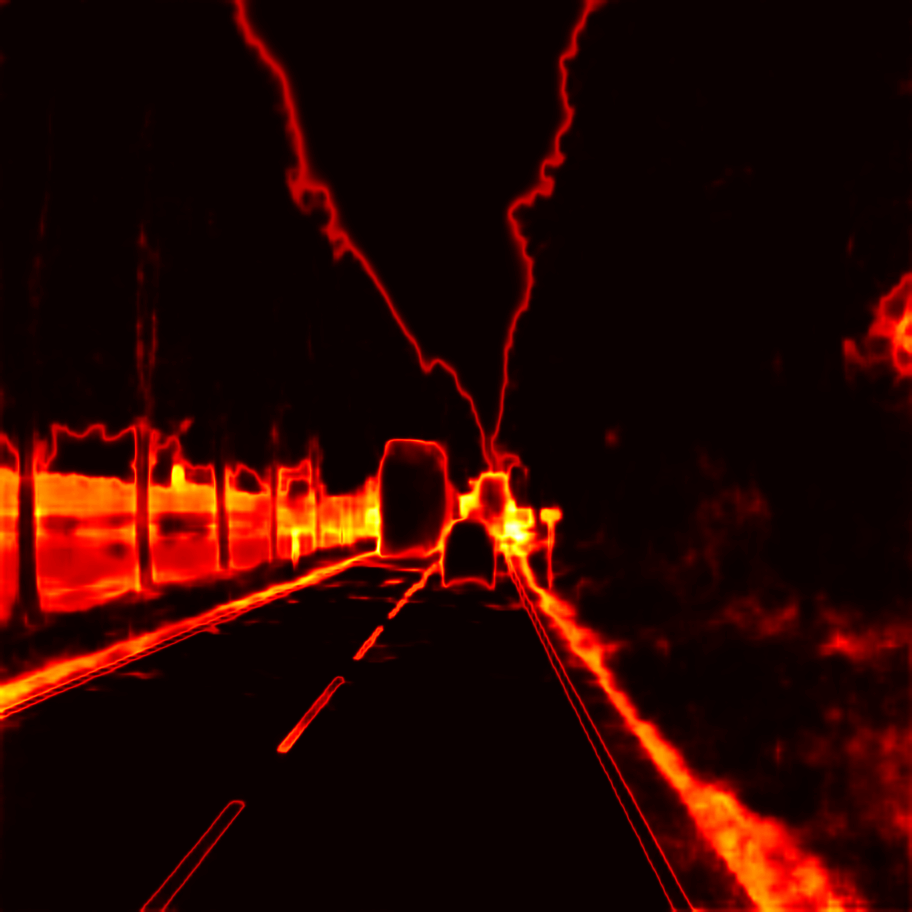} &
      \includegraphics[width=0.1\textwidth]{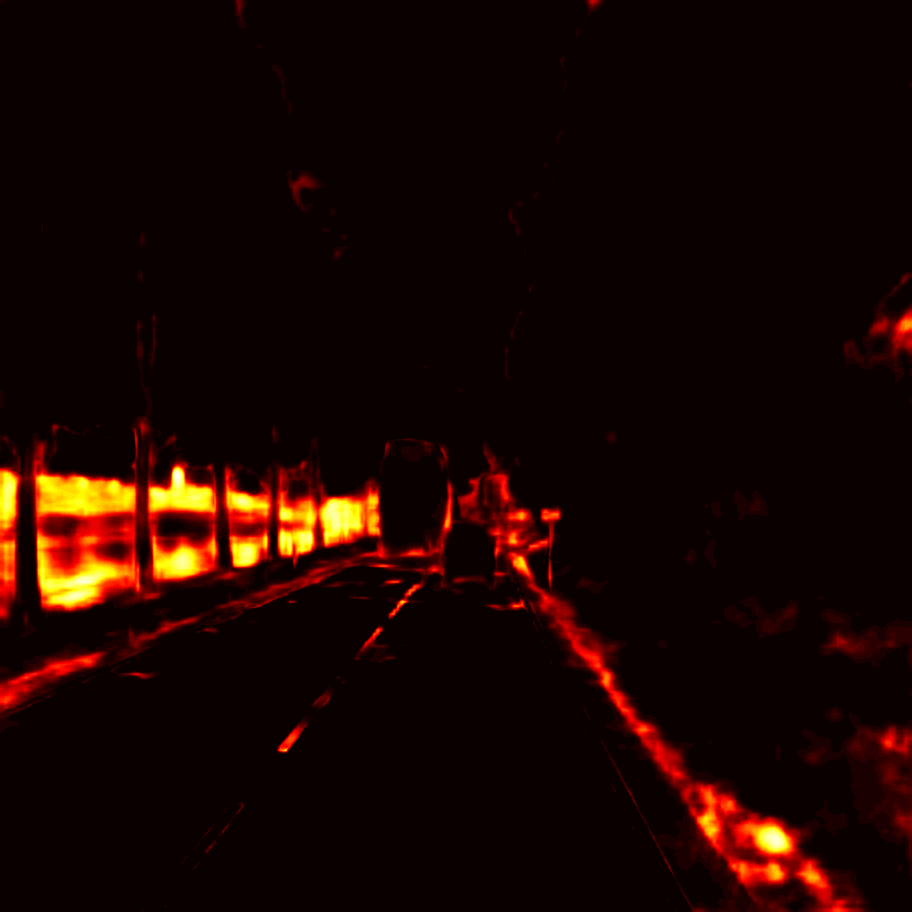} &
      \includegraphics[width=0.1\textwidth]{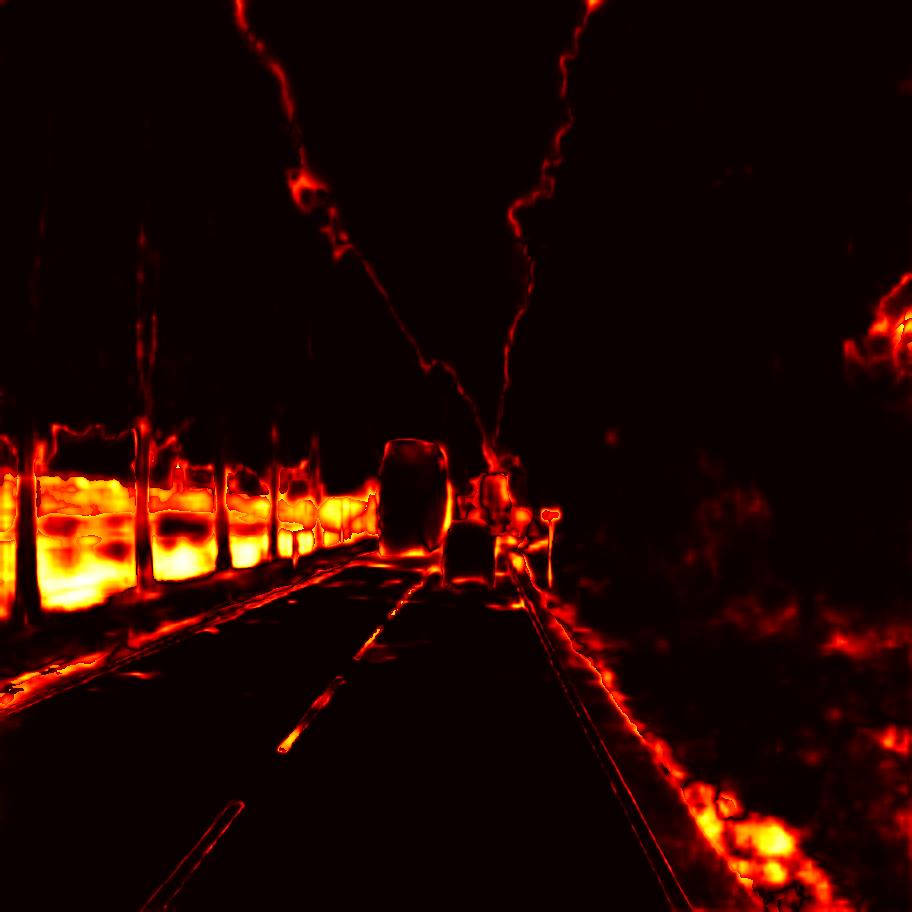} 
      &  \\

  \end{tabular}%
}
\end{table*}

\subsection{Model Performance and Calibration}
All MoE models outperform the baselines and the ensemble regarding semantic segmentation accuracy (see Figure~\ref{fig:miou}), with the classwise gate leading to higher mIoU values. We then evaluate the calibration of the uncertainty estimates extracted with different methods from the baseline, the ensemble, and four different MoE architectures on in-distribution (highway and urban) and out-of-distribution (ambiguous) test data (see Figure~\ref{fig:uq_comparison}). MoEs outperform baselines and ensembles on OOD data in all metrics apart from MCE, highlighting their potential as strong uncertainty estimators beyond just segmentation accuracy. This improved calibration is most consistently achieved when using PE as the uncertainty estimator, while MI performs less reliably and EV consistently underperforms across both domains. The differences between weighted and stacked approaches to computing PE and MI are relatively minor. Moreover, classwise gating (with and without conv) shows slightly better calibration than simple gating, particularly for MCE and NLL, suggesting that more flexible or fine-grained gating improves confidence alignment. The weighted uncertainty formulation is, however, only applicable to simple gates, so no comparison over all four MoE architectures is possible. In summary, the analysis of the calibration metrics shows that MoEs can achieve well-calibrated uncertainty estimates, surpassing standard ensembles on OOD data. 

The analysis of the gate entropy (see Figure~\ref{fig:gate-entropy}) demonstrates that the gating mechanism shows almost no difference between ID and OOD inputs, indicating a lack of distributional awareness. Due to simpler routing decisions, simple gates are better calibrated than classwise gates. Also, adding a convolutional layer, while improving the segmentation accuracy,  has not reduced routing uncertainty for the classwise gate. 

Visual analysis of the resulting uncertainty (see Table ~\ref{tab:uq_examples}) shows that a higher uncertainty is observed for all methods along object boundaries, such as the edges of the trees and road markings for all approaches, implying aleatoric uncertainty. Also, highly uncertain areas often coincide with regions with higher prediction errors. For MoEs, higher uncertainty is additionally present in the areas near the highway border, implying epistemic uncertainty.




\subsection{Robustness of Uncertainty Quantification under Data Shift}

In addition to the calibration analysis, we evaluate model robustness under data shift to assess the performance and reliability of the proposed uncertainty quantification methods under varying degrees of input corruption. Similarly to~\cite{ovadia2019can}, we use a total of 19 image corruptions, including 15 standard corruptions used by Hendrycks et al.~\cite{hendrycks2019benchmarking}, along with four additional types: speckle noise, Gaussian blur, spatter, and saturation. Because MoE with a classwise gate without an additional convolutional layer consistently showed strong calibration on OOD data in our experiments (see Figure~\ref{fig:uq_comparison}), we use it for the data shift experiments. The evaluation was performed on the combined \textit{highway-urban} A2D2 test data. 

\begin{figure}[t]
  \centering
  \begin{subfigure}[t]{\linewidth}
    \centering
    \includegraphics[width=\linewidth]{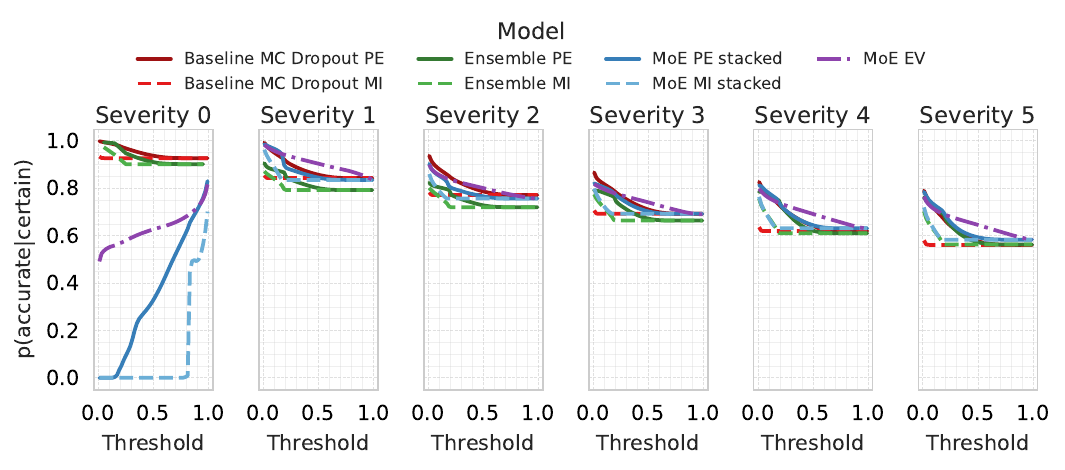}
    \label{fig:a2d2_moe_pe_accurate_certain}
  \end{subfigure}

  \begin{subfigure}[t]{\linewidth}
    \centering
    \includegraphics[width=\linewidth]{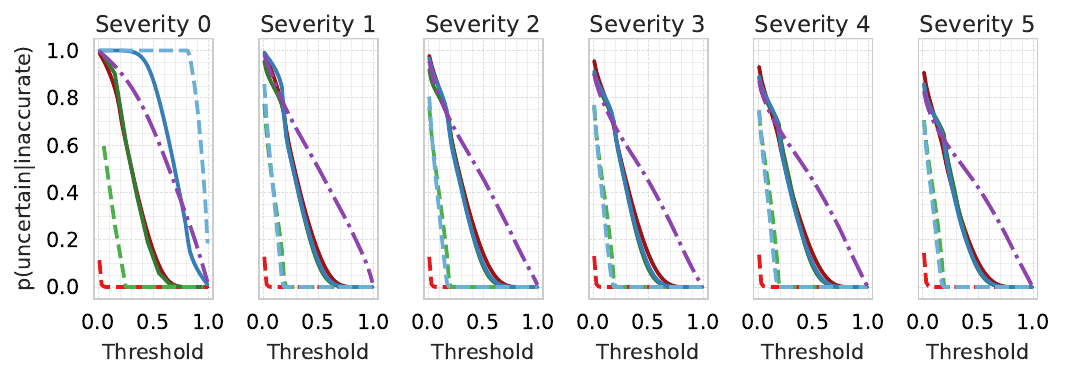}
    \label{fig:a2d2_moe_pe_uncertain_inaccurate}
  \end{subfigure}

  \begin{subfigure}[t]{\linewidth}
    \centering
    \includegraphics[width=\linewidth]{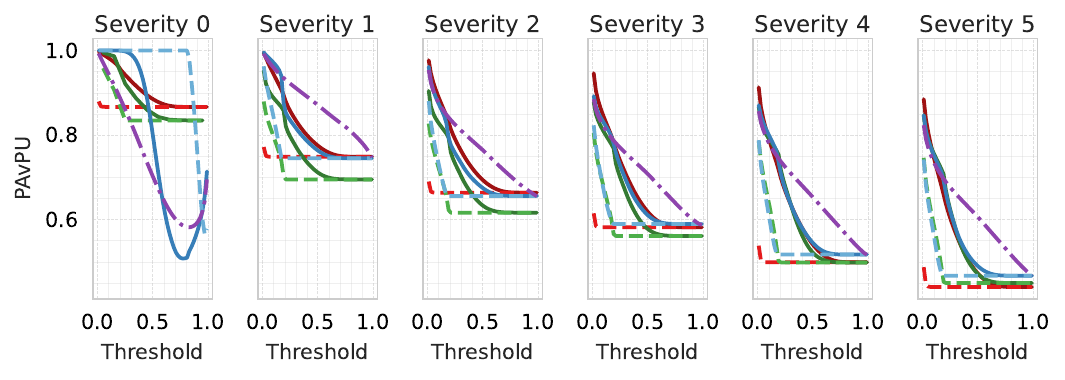}
    \label{fig:a2d2_moe_pavpu}
  \end{subfigure}

  \caption{Conditional correctness metrics for the A2D2 models under data shift. Higher values indicate better alignment between the model's uncertainty and its prediction accuracy.}
  \label{fig:comparison-a2d2-robustness}
\end{figure}

In terms of the conditional correctness metrics $p$(uncertain$\vert$inaccurate), $p$(accurate$\vert$certain), and PAvPU (see Figure~\ref{fig:comparison-a2d2-robustness}), MoEs show more reliable uncertainty estimates than baselines and ensembles. They maintain higher accuracy among confident predictions, better detection of uncertain errors, and superior overall uncertainty-performance balance as measured by PAvPU. Among different MoE uncertainty estimation methods, EV performs the best over all severity levels.

Furthermore, Table \ref{tab:uq_peak_values_grouped} confirms that MoEs achieve the highest peak values for the conditional metrics and the area under the PAvPU curve (AU–PAvPU) score. MoEs with MI also yield the highest AU-PAvPU, indicating consistently strong performance across thresholds. In contrast, ensembles and baselines show lower peak and area metrics. Note that a 100\,\% peak occurs when there are very few inaccurate or very few certain pixels at the respective threshold.

\begin{figure}[t]
\begin{subfigure}[t]{\columnwidth}
        \centering
        \includegraphics[width=0.8\textwidth]{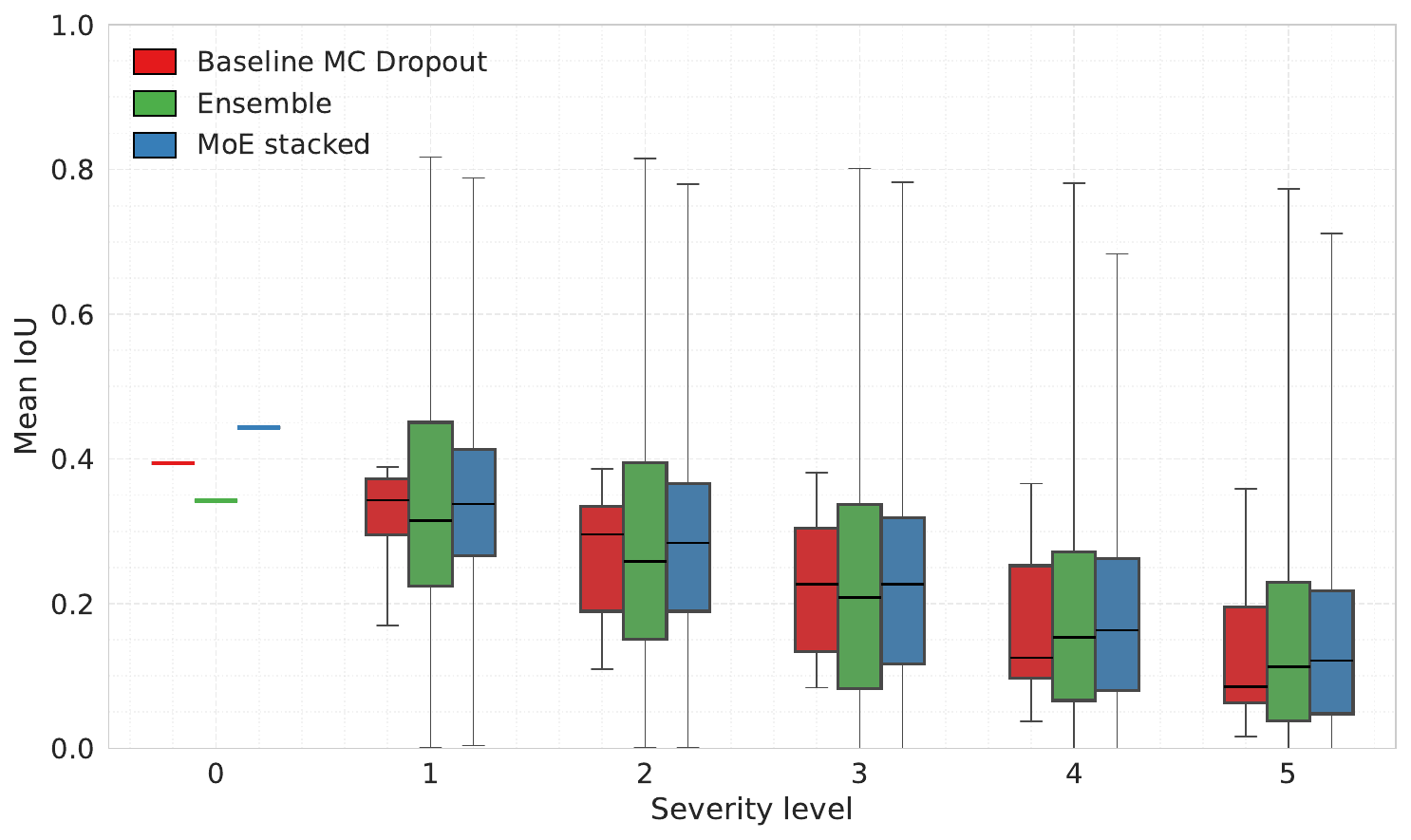}
        \label{fig:robustness-comparison-miou}
    \end{subfigure}
     \begin{subfigure}[t]{\columnwidth}
        \centering
        \includegraphics[width=0.8\textwidth]{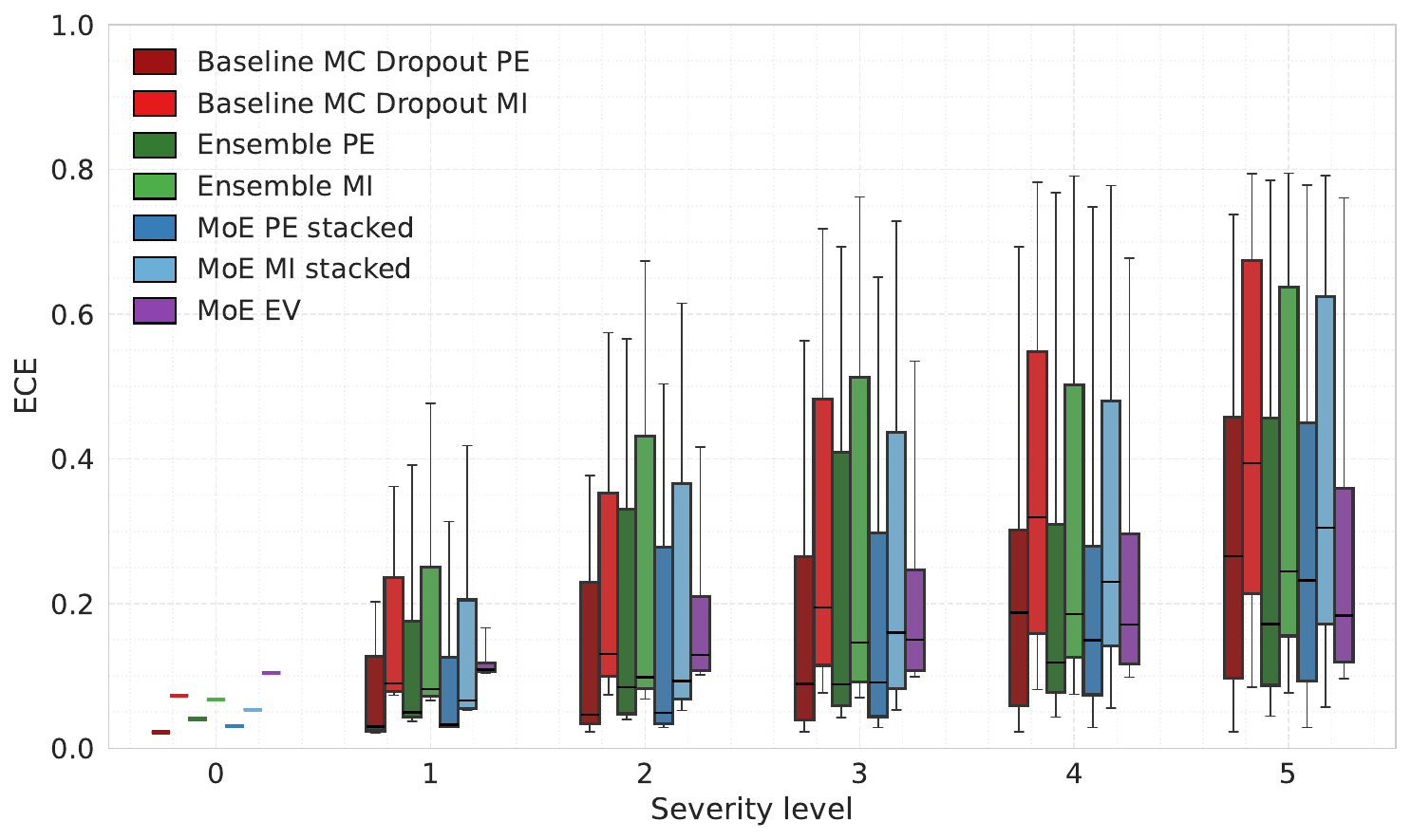}
        \label{fig:robustness-comparison-a2d2-ece}
    \end{subfigure}
    \begin{subfigure}[t]{\columnwidth}
        \centering
        \includegraphics[width=0.8\textwidth]{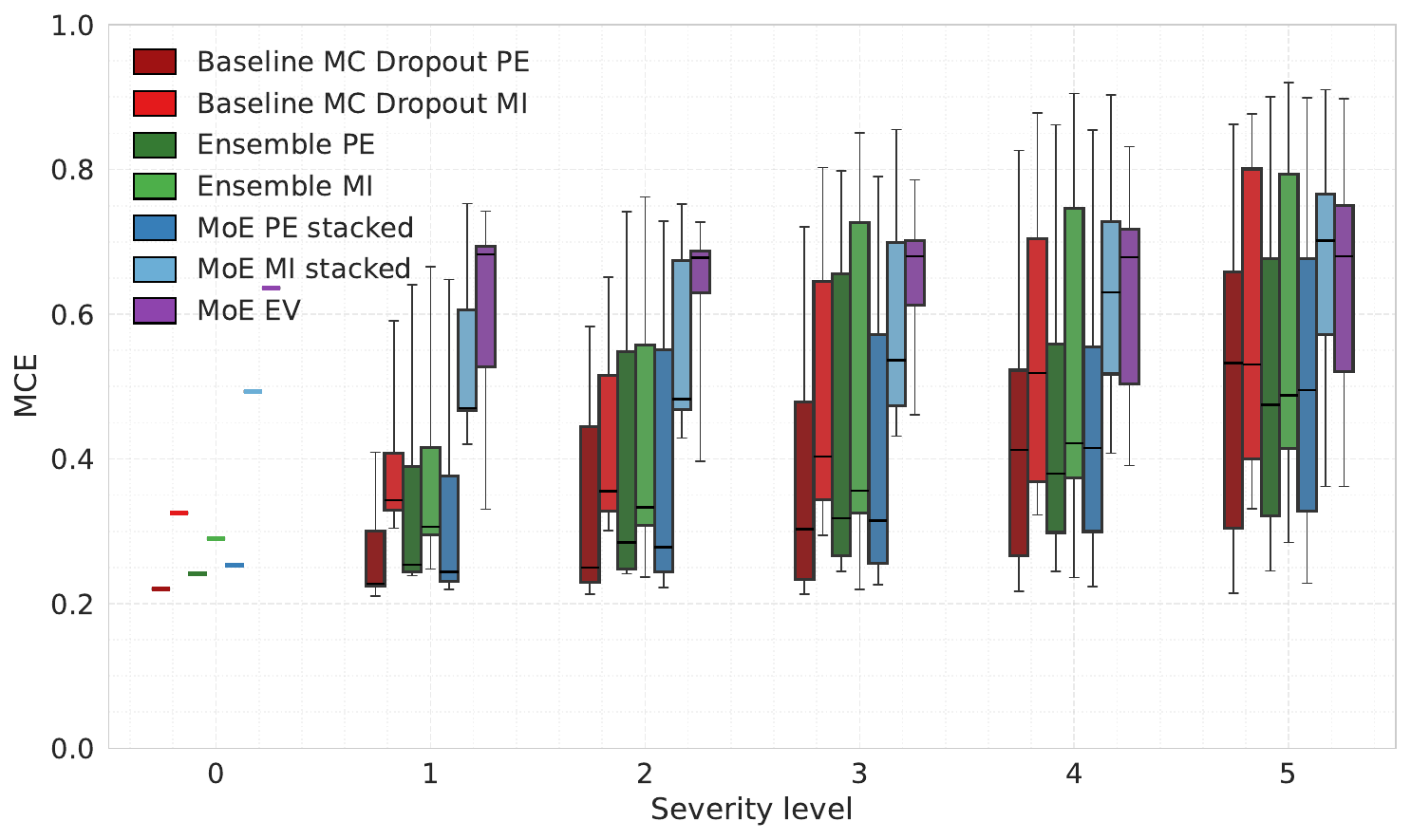}
        \label{fig:robustness-comparison-a2d2-mce}
    \end{subfigure}
    \begin{subfigure}[t]{\columnwidth}
        \centering
        \includegraphics[width=0.8\textwidth]{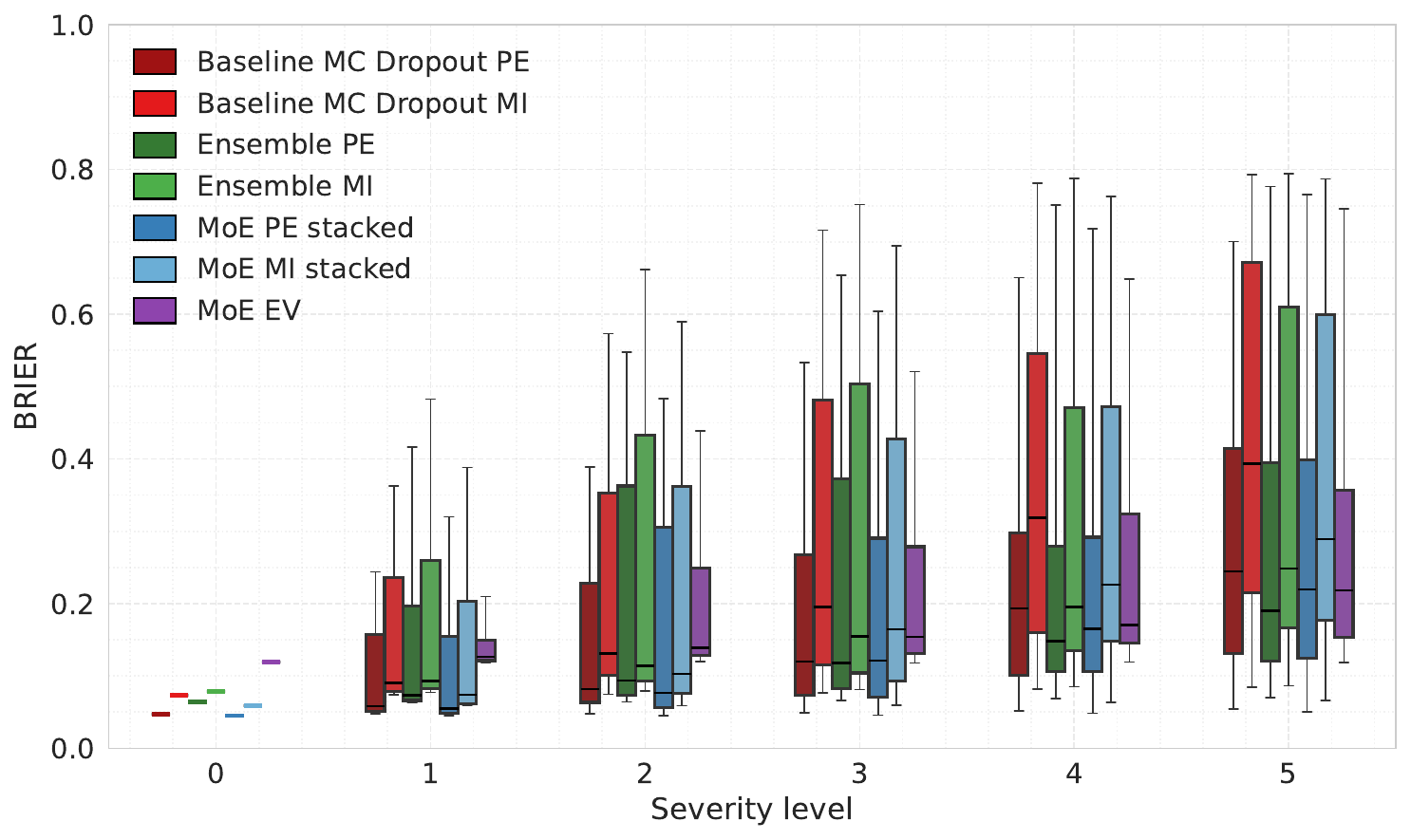}
        \label{fig:robustness-comparison-a2d2-brier}
    \end{subfigure}
    \begin{subfigure}[t]{\columnwidth}
        \centering
        \includegraphics[width=0.8\textwidth]{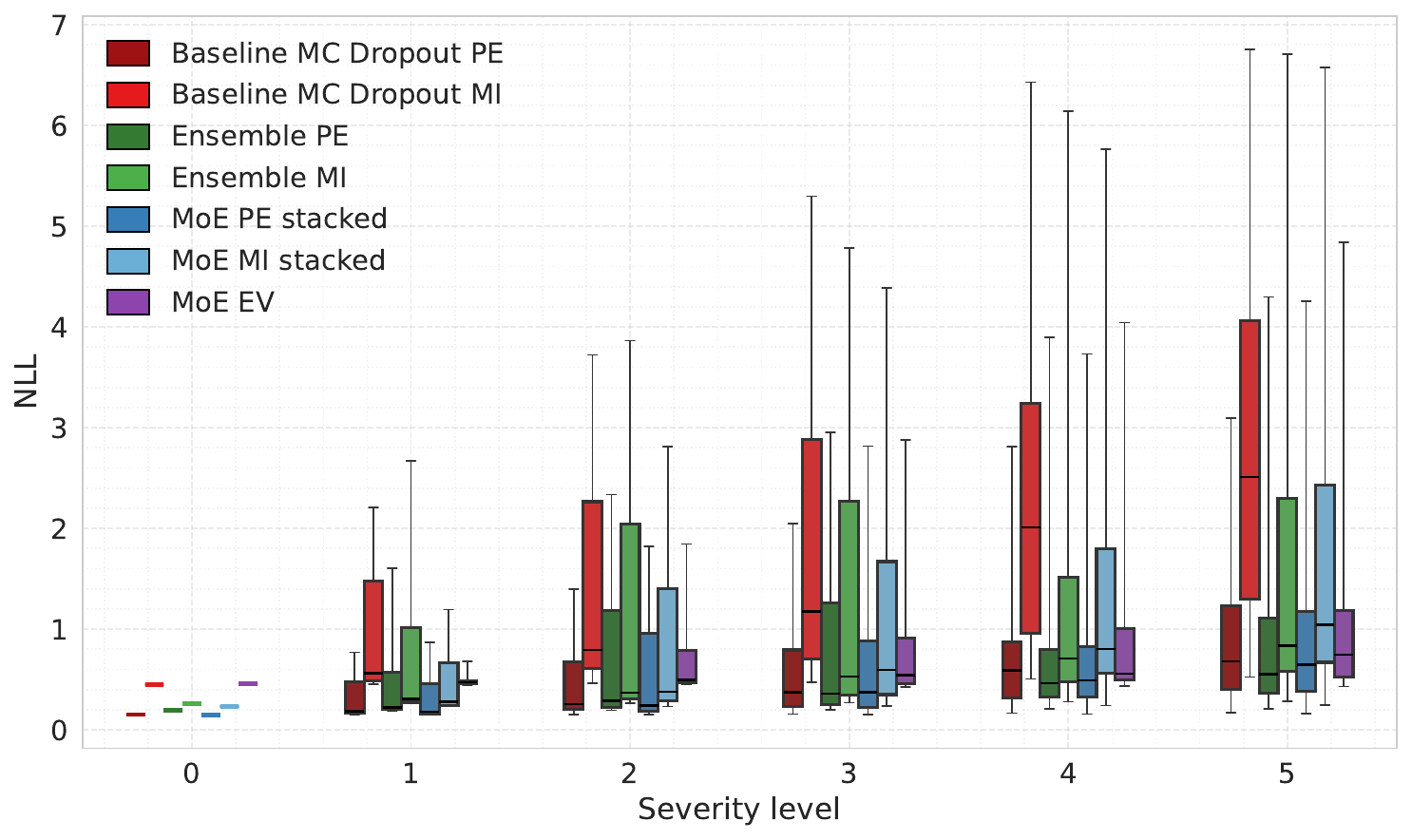}
        \label{fig:robustness-comparison-a2d2-nll}
    \end{subfigure}
    \caption{Uncertainty calibration of A2D2 models under data shift. Lower values indicate better-calibrated uncertainty estimates.}
    \label{fig:robustness-comparison}
\end{figure}

The calibration metrics demonstrate a different behavior (see Figure~\ref{fig:robustness-comparison}). While MoEs with PE provide strong and consistent calibration under data shift, ensembles with PE often achieve lower calibration errors, particularly in ECE and Brier score. MoE variants, e.g., PE-stacked, remain competitive but do not consistently outperform ensembles.

\begin{table}[t]
\centering
\caption{Peak conditional‐probability metrics ($\max p(\mathrm{a|c})$ is $\max p(\text{accurate}\mid\text{certain})$, and 
$\max p(\mathrm{u|i})$ is $\max p(\text{uncertain}\mid\text{inaccurate})$) and AU-PAvPU in \%. }
\label{tab:uq_peak_values_grouped}
\resizebox{1.0\linewidth}{!}{
\begin{tabular}{|r | l |c c c c|}
\hline
\textbf{Method} & \textbf{Model} 
    & \textbf{$\max$} 
    & \textbf{$\max$ }
    & \textbf{$\max$ }
    & \textbf{$AU-$} \\ 
    
&  
    & \textbf{$p(\mathrm{a|c})\,\uparrow$ }
    & \textbf{$p(\mathrm{u|i})\,\uparrow$} 
    & \textbf{$\mathrm{PAvPU}\,\uparrow$} 
    & \textbf{$\mathrm{PAvPU}\,\uparrow$} \\
\hline
PE    & Baseline (MC Dropout)   & \textbf{99.74} & 99.08 & 99.71 & \textbf{87.56} \\
    & Ensemble                & 99.28 & 97.61 & 99.49 & 79.50 \\
    & MoE                      & 98.48 & \textbf{100.00}\textsuperscript{*} & \textbf{100.00}\textsuperscript{*} & 78.95 \\
\hline
MI    & Baseline (MC Dropout)   & 93.30 & 14.39 & 87.93 & 84.80 \\
    & Ensemble                & \textbf{97.24} & 77.43 & 95.83 & 78.00 \\
    & MoE                      & 96.18 & \textbf{100.00}\textsuperscript{*} & \textbf{100.00}\textsuperscript{*} & \textbf{93.18} \\
\hline
EV & MoE                     & 98.20 & 99.58 & 99.31 & 71.94 \\
\hline
\end{tabular}
}
\end{table}

This discrepancy over metrics arises from their properties. Calibration metrics assess how well predicted probabilities reflect true likelihoods, focusing on global alignment between confidence and accuracy. In contrast, conditional metrics evaluate how effectively uncertainty distinguishes between correct and incorrect predictions, thus capturing the utility of uncertainty for decision-making. MoEs tend to perform better on conditional metrics because their uncertainty estimates more reliably identify uncertain or erroneous predictions, even if their overall confidence calibration is less accurate than that of ensembles.

\section{Evaluation of MoEs with Semantically Overlapping Experts}
The MoE architecture used for the experiments in Section~\ref{sec:disjoint} assumes two experts trained on semantically disjoint subsets of data. Due to the selected semantic split (highway-urban), we could not evaluate the impact of the number of experts on the quality of the extracted uncertainty estimates. Therefore, we additionally perform experiments with semantically overlapping experts on the Cityscapes data. 

\subsection{Experimental Setup}

\textbf{Model and Dataset}: We use the same MoE architecture as above with \texttt{DeepLabv3+} as an expert and the Cityscapes dataset~\cite{cordts2016cityscapes} with images resized to $512\times1024$ pixels. 

\textbf{Training:}  Each expert was pre-trained on 20K coarsely-annotated images for 50 epochs and then fine-tuned on 2K fine-annotated images for 100 epochs with a batch size of two. Each expert was fine-tuned with a different weight initialization to ensure variability in the segmentation accuracy. The MoEs with a simple gate and a convolutional layer were trained for 100 epochs with a batch size of eight. Other settings are the same as in Section~\ref{sec:disjoint}.

\subsection{Model Performance and Calibration}
Table \ref{tab:cityscapes-results} shows the resulting segmentation performance of the experts and MoE. As evident by the values, the experts did not specialize in a domain, and the MoEs did not gain any segmentation performance by adding experts. The results also show that scaling the number of experts in MoEs slightly improves the NLL, with the 10-expert MoE achieving the best NLL score among the MoEs for both PE and EV. However, other calibration metrics (ECE, MCE, Brier) do not improve consistently, particularly for EV, where MCE remains substantially higher than ensembles, even with 10 experts. Compared to ensembles, MoEs have slightly worse mIoU and worse calibration metrics overall. Interestingly, ensembles benefit more from increasing the number of experts in terms of accuracy and calibration, while MoEs plateau or even degrade slightly in some metrics. This suggests that while scaling up MoEs can marginally enhance uncertainty estimation, as is the case for NLL, it does not straightforwardly improve calibration, and may require more sophisticated gating or regularization strategies to leverage additional experts fully.

\begin{table}[t]
\centering
\caption{Performance and calibration evaluation on the Cityscapes test data. The ECE, MCE, and Brier score values are in the [0,1] interval. NLL has an unbounded logarithmic scale.}
\label{tab:cityscapes-results}    
\resizebox{1.0\linewidth}{!}{
    \begin{tabular}{|r|l|ccccc|}
    \hline
    \textbf{Method} & \textbf{Model} & \textbf{mIoU $\uparrow$} & \textbf{ECE $\downarrow$} & \textbf{MCE $\downarrow$} & \textbf{Brier $\downarrow$} & \textbf{NLL $\downarrow$} \\
    \hline
    & Baseline (MC dropout)           & \textbf{0.641} & 0.018 & 0.222 & 0.049 & 0.160 \\
    & Ensemble, 2 experts             &        0.640   & 0.017 & 0.208 & 0.048 & 0.156 \\
    & Ensemble, 5 experts             &        0.640   & \textbf{0.017} & \textbf{0.207} & \textbf{0.048} & \textbf{0.156} \\
    & Ensemble, 10 experts            &        0.639   & 0.018 & 0.215 & 0.048 & 0.156 \\ 
PE  & MoE stacked, 2 experts          &        0.628   & 0.019 & 0.228 & 0.049 & 0.160 \\
    & MoE stacked, 5 experts          &        0.627   & 0.020 & 0.234 & 0.050 & 0.162 \\
    & MoE stacked, 10 experts         &        0.635   & 0.019 & 0.233 & \textbf{0.048} & 0.157 \\
    & MoE weighted, 2 experts         &        0.628   & 0.019 & 0.240 & 0.049 & 0.159 \\
    & MoE weighted, 5 experts         &        0.627   & 0.020 & 0.235 & 0.050 & 0.163 \\
    & MoE weighted, 10 experts        &        0.635   & 0.019 & 0.229 & \textbf{0.048} & 0.157 \\
    \hline
MI  & Baseline (MC dropout)           &        0.641   & 0.071 & 0.391 & 0.072 & \textbf{0.391} \\
    & Ensemble, 2 experts             &        0.640   & 0.071 & 0.071 & 0.071 & 0.588 \\
    & Ensemble, 5 experts             &        0.640   & 0.071 & \textbf{0.071} & 0.071 & 0.504 \\
    & Ensemble, 10 experts            &        0.639   & 0.070 & 0.078 & \textbf{0.070} & 0.423 \\
    & MoE stacked, 2 experts          &        0.628   & 0.074 & 0.101 & 0.074 & 0.448 \\
    & MoE weighted, 2 experts         &        0.628   & 0.075 & 0.075 & 0.075 & 0.876 \\
    & MoE stacked, 5 experts          &        0.627   & 0.075 & 0.085 & 0.076 & 0.453 \\
    & MoE weighted, 5 experts         &        0.627   & 0.076 & 0.076 & 0.076 & 0.874 \\
    & MoE stacked, 10 experts         &        0.635   & 0.071 & 0.090 & 0.072 & 0.459 \\
    & MoE weighted, 10 experts        &        0.635   & \textbf{0.069} & 0.395 & 0.071 & 0.673 \\
    \hline
EV  & MoE stacked, 2 experts          &        0.628   & 0.047 & 0.434 & 0.061 & 0.217 \\
    & MoE stacked, 5 experts          &        0.627   & 0.044 & \textbf{0.425} & 0.060 & 0.205 \\
    & MoE stacked, 10 experts         &        0.635   & \textbf{0.040} & 0.496 & \textbf{0.057} & \textbf{0.188} \\
    \hline
    \end{tabular}
}
\end{table}

\section{Conclusion}
Previous works have demonstrated that MoE models can outperform single models and ensembles in terms of segmentation performance. In this work, we investigated uncertainty quantification in MoE architectures for semantic segmentation, focusing on extracting predictive entropy and mutual information and proposing a novel expert variance measure. Without requiring architectural modifications, we showed that MoEs can produce competitive and often superior uncertainty estimates compared to standard ensembles and MC dropout, particularly under dataset shift. 

In experiments with two semantically disjoint experts on the A2D2 dataset, MoEs demonstrated better calibration in conditional probability metrics and comparable or slightly worse performance in standard calibration scores such as ECE and Brier on OOD data. Our analysis also showed that MoEs maintain more consistent PAvPU performance across increasing severity levels of input corruption. 

Scaling the number of experts to five and ten using the Cityscapes dataset revealed modest improvements in calibration metrics, particularly in NLL, suggesting that increasing model capacity can enhance uncertainty estimation. 

Furthermore, the separation of predictive and routing uncertainty components, for example, via predictive entropy and gate entropy evaluated in this work, offers potential for disentangling epistemic and aleatoric uncertainty. 

To the best of our knowledge, our work is the first to demonstrate that reliable uncertainty estimates can be obtained from standard, unmodified MoEs without requiring any architectural changes or explicit uncertainty modeling. Future work can explore adaptive expert selection, deeper gating strategies, and richer expert diversity to improve uncertainty robustness and interoperability. Finally, the superior results observed on OOD data suggest that MoEs can serve as an OOD or anomaly detection method. 

\section*{Acknowledgment}

This work was supported by funding from the Topic Engineering Secure Systems of the Helmholtz Association (HGF) and by KASTEL Security Research Labs (46.23.03).

{
    \small
    \bibliographystyle{ieeenat_fullname}
    \bibliography{references}
}

\end{document}